\documentclass[10pt,twocolumn,letterpaper]{article}

\usepackage{iccv}
\usepackage{times}
\usepackage{subfig}
\usepackage{multirow}
\usepackage{epsfig}
\usepackage{graphicx}
\usepackage{amsmath}
\usepackage{amssymb}
\usepackage[outercaption]{sidecap}  

\usepackage[pagebackref=true,breaklinks=true,letterpaper=true,colorlinks,bookmarks=false]{hyperref}

\iccvfinalcopy 
\def\iccvPaperID{52} 

\title{HOGgles: Visualizing Object Detection Features\thanks{This is a pre-print of our conference paper. We made it publicly available early in the hopes others find it useful. Last modified \today.}}
\title{Inverting and Visualizing Features for Object Detection\thanks{This is a pre-print of our conference paper. We made it publicly available early in the hope others find it useful. Last modified \today.}}
\author{Carl Vondrick, Aditya Khosla, Tomasz Malisiewicz, Antonio Torralba\\
Massachusetts Institute of Technology\\
\texttt{\{vondrick,khosla,tomasz,torralba\}@csail.mit.edu}}

\DeclareMathOperator*{\argmax}{argmax}
\DeclareMathOperator*{\argmin}{argmin}

\begin{document}

\maketitle

%
%

\begin{abstract}

We introduce algorithms to visualize feature spaces used by object detectors.
The tools in this paper allow a human to put on `HOG goggles' and perceive the
visual world as a HOG based object detector sees it. We found that these
visualizations allow us to analyze object detection systems in new ways and
gain new insight into the detector's failures. For example, when we visualize
the features for high scoring false alarms, we discovered that, although they
are clearly wrong in image space, they do look deceptively similar to true
positives in feature space. This result suggests that many of these false
alarms are caused by our choice of feature space, and indicates that creating a
better learning algorithm or building bigger datasets is unlikely to correct
these errors. By visualizing feature spaces, we can gain a more intuitive
understanding of our detection systems.

\end{abstract}

\section{Introduction}

\begin{figure}
\centering

\includegraphics[width=0.8\linewidth]{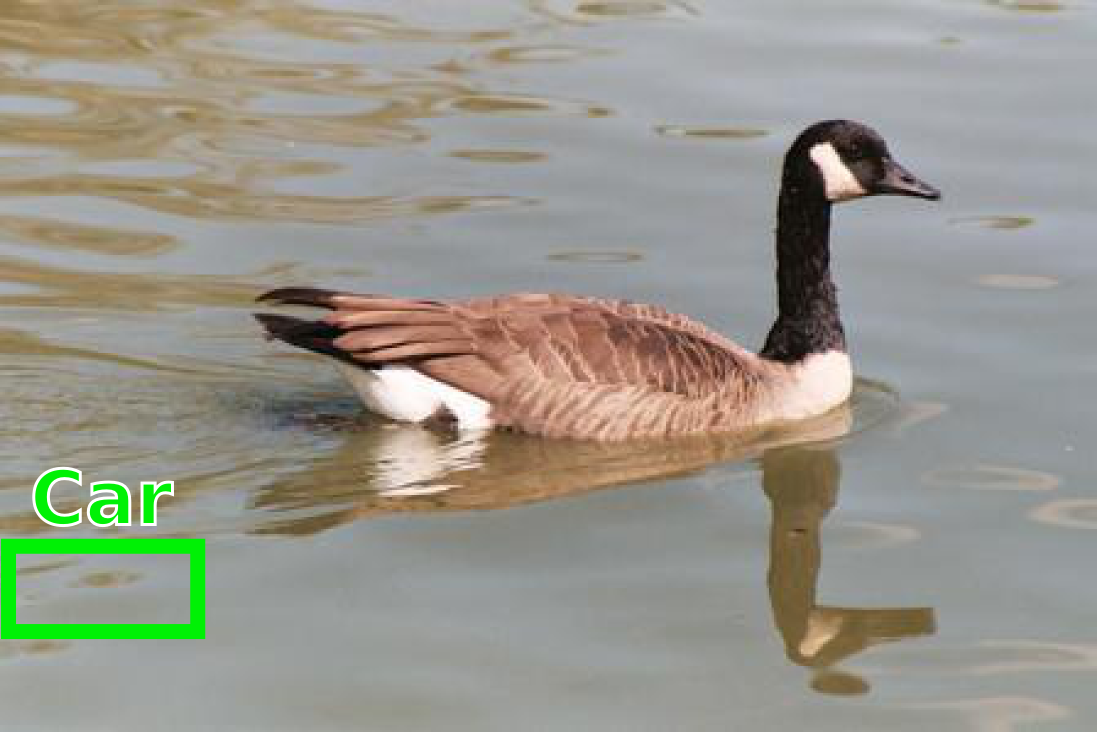}

\caption{An image from PASCAL and a high scoring car detection from
DPM \cite{felzenszwalb2010object}. Why did the detector fail?}

\label{fig:teaser}

\includegraphics[width=\linewidth]{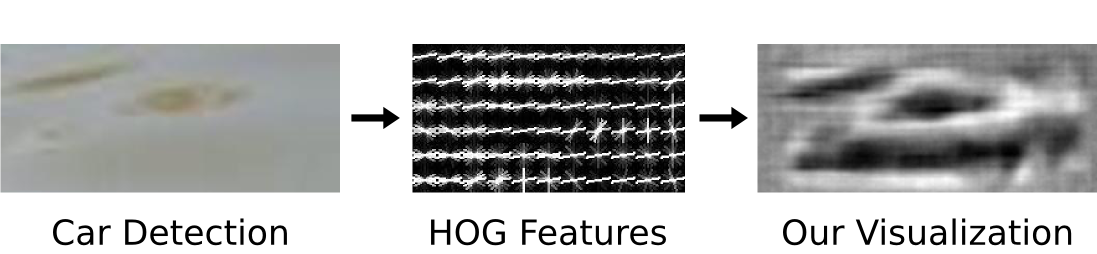}

\caption{We show the crop for the false car detection from
Figure \ref{fig:teaser}. On the right, we show our visualization of the HOG
features for the same patch. Our visualization reveals that this false alarm
actually looks like a car in HOG space.}

\label{fig:teaser2}

\vspace{-1.5em}

\end{figure}

Figure \ref{fig:teaser} shows a high scoring detection from an object detector
with HOG features and a linear SVM classifier trained on PASCAL.  Despite our
field's incredible progress in object recognition over the last decade,
\emph{why} do our detectors still think that lakes look like cars?

\begin{figure*}
\centering

\includegraphics[width=\linewidth]{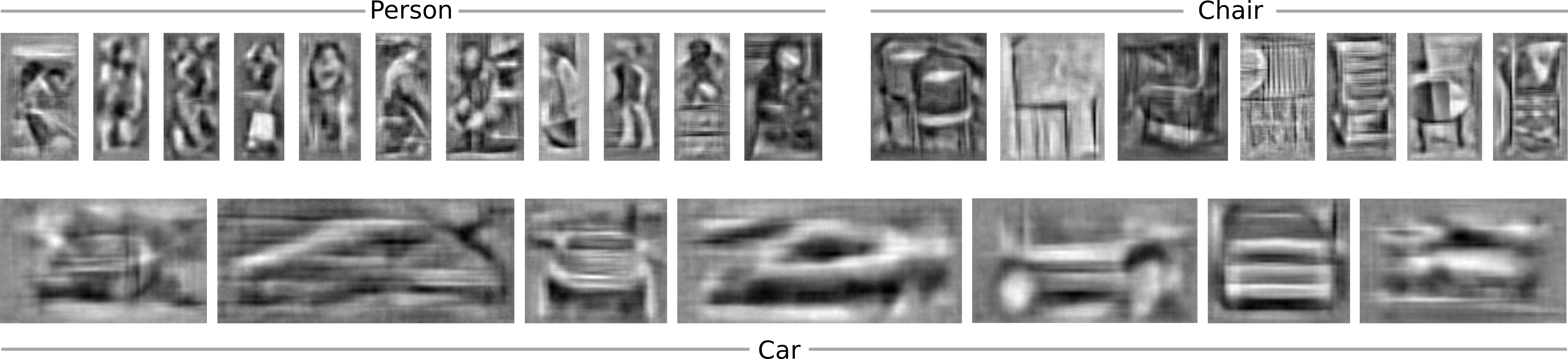}

\caption{We visualize some high scoring detections from the deformable parts
model \cite{felzenszwalb2010object} for person (top left), chair (top right),
and car (bottom). Can you guess which are false alarms? Take a minute to study
this figure, then see Figure \ref{fig:topdetsrgb} for the corresponding RGB
patches.}

\label{fig:topdets}

\vspace{-1em}

\end{figure*}

Unfortunately, computer vision researchers are often unable to explain the
failures of object detection systems. Some researchers blame the features,
others the training set, and even more the learning algorithm. Yet, if we wish
to build the next generation of object detectors, it seems crucial to
understand the failures of our current detectors.

In this paper, we introduce tools to explain many of the failures of object
detection systems.\footnote{Code is available online at
\url{http://mit.edu/vondrick/ihog}} We present algorithms to visualize the
feature spaces of object detectors. Since features are too high dimensional for
humans to directly inspect, our visualization algorithms work by inverting
features back to natural images.  We found that these inversions provide an
intuitive and accurate visualization of the feature spaces used by object
detectors.

Figure \ref{fig:teaser2} shows the output from our visualization on the
features for the false car detection. This visualization reveals that,
while there are clearly no cars in the original image, there is a car hiding
in the HOG descriptor. HOG features see a slightly different visual world than
what we see, and by visualizing this space, we can gain a more intuitive
understanding of our object detectors. 

Figure \ref{fig:topdets} inverts more top detections on PASCAL for a few
categories. Can you guess which are false alarms? Take a minute to study the
figure since the next sentence might ruin the surprise. Although every
visualization looks like a true positive, all of these detections are actually
false alarms.  Consequently, we can conclude that, even with a better learning
algorithm or more data, these false alarms will likely persist. In other words,
\emph{the features are to blame}.

\begin{figure}
\includegraphics[width=\linewidth]{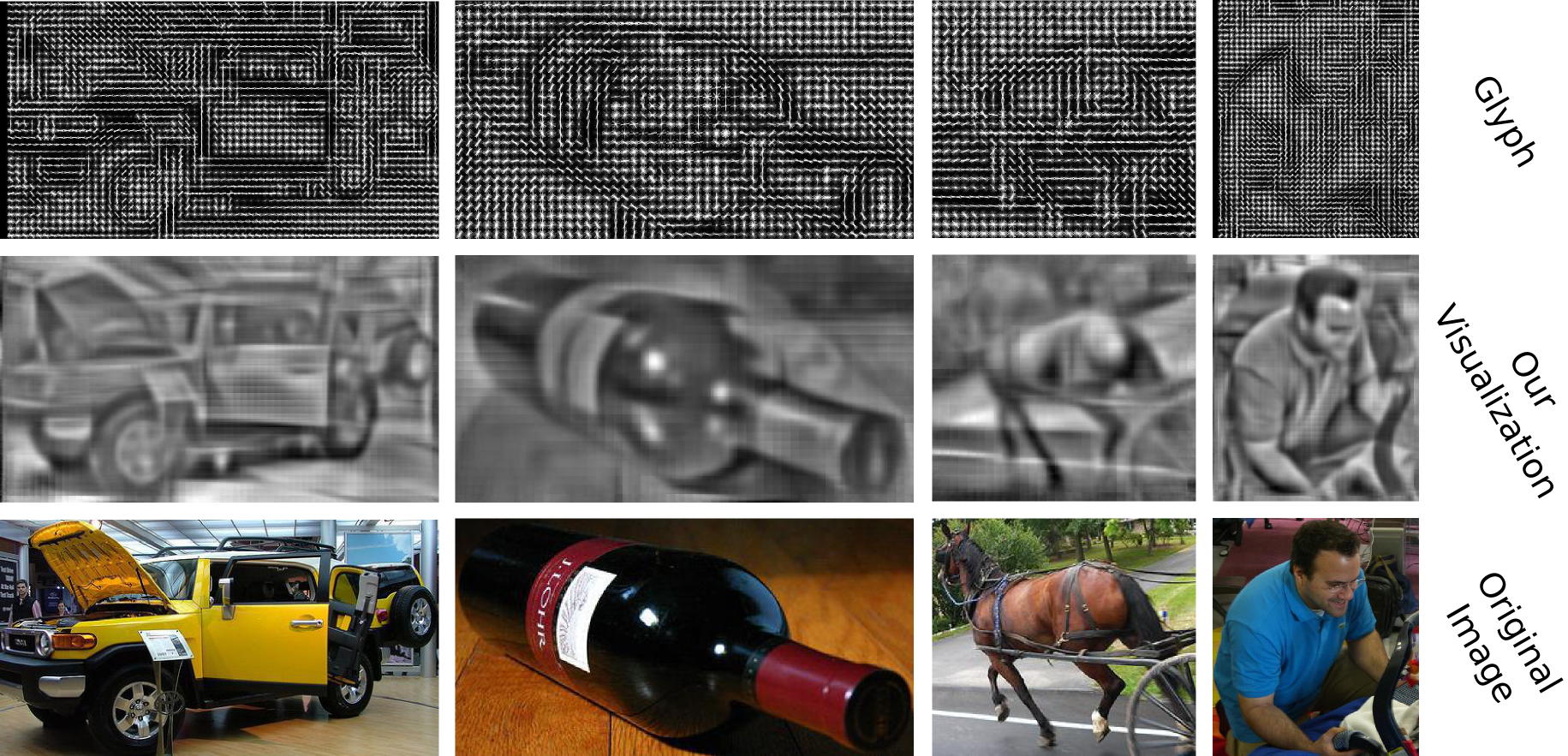}
\caption{Examples of our HOG feature visualization.}
\label{fig:example}
\vspace{-1em}
\end{figure}

The principle contribution of this paper is the presentation of algorithms for
visualizing features used in object detection. To this end, we present four
algorithms to invert object detection features to natural images. Although we
focus on HOG features in this paper, our approach is general and can be applied
to other features as well. We evaluate our inversions with both automatic
benchmarks and a large human study, and we found our visualizations are
perceptually more accurate at representing the content of a HOG feature than
existing methods; see Figure \ref{fig:example} for a comparison between our
visualization and HOG glyphs. We then use our visualizations to inspect the
behaviors of object detection systems and analyze their features. Since we hope
our visualizations will be useful to other researchers, our final contribution
is a public feature visualization toolbox. 

\section{Related Work}

Our visualization algorithms extend an actively growing body of work in
feature inversion.  Torralba and Oliva, in early work \cite{torralba2002depth},
described a simple iterative procedure to recover images only given gist
descriptors \cite{oliva2006building}. Weinzaepfel et al.\
\cite{weinzaepfel2011reconstructing} were the first to reconstruct an image
given its keypoint SIFT descriptors \cite{lowe1999object}. Their approach
obtains compelling reconstructions using a nearest neighbor based approach on a
massive database. d'Angelo et al.\ \cite{d2012beyond} then developed an
algorithm to reconstruct images given only LBP features
\cite{calonder2010brief,alahi2012freak}. Their method analytically solves for
the inverse image and does not require a dataset.

While \cite{weinzaepfel2011reconstructing,d2012beyond,torralba2002depth} do a
good job at reconstructing images from SIFT, LBP, and gist features, our
visualization algorithms have several advantages. Firstly, while existing
methods are designed for specific features, our visualization algorithms we
propose are feature independent. Since we cast feature inversion as a machine
learning problem, our algorithms can be used to visualize any feature. In this
paper, we focus on features for object detection, the most popular of which is
HOG. Secondly, our algorithms are fast: our best algorithm can invert features
in under a second on a desktop computer, enabling interactive visualization.
Finally, to our knowledge, this paper is the first to invert HOG.

Our visualizations enable analysis that complement a recent line of papers that
provide tools to diagnose object recognition systems, which we briefly review
here.  Parikh and Zitnick \cite{parikh2011human,parikh2010role} introduced a
new paradigm for human debugging of object detectors, an idea that we adopt in
our experiments.  Hoiem et al.\ \cite{hoiem2012diagnosing} performed a large
study analyzing the errors that object detectors make. Divvala et al.\
\cite{divvala2012important} analyze part-based detectors to determine which
components of object detection systems have the most impact on performance.  Tatu et
al.\ \cite{tatu2011exploring} explored the set of images that generate
identical HOG descriptors. Liu and Wang \cite{liu2012has} designed algorithms
to highlight which image regions contribute the most to a classifier's
confidence. Zhu et al.\ \cite{zhuwe} try to determine whether we have reached
Bayes risk for HOG.  The tools in this paper enable an alternative mode to
analyze object detectors through visualizations.  By putting on `HOG glasses'
and visualizing the world according to the features, we are able to gain a better
understanding of the failures and behaviors of our object detection systems.



\section{Feature Visualization Algorithms}

We pose the feature visualization problem as one of feature inversion, i.e.\
recovering the natural image that generated a feature vector. Let $x \in
\mathbb{R}^{D}$ be an image and  $y = \phi(x)$ be the corresponding HOG feature
descriptor. Since $\phi(\cdot)$ is a many-to-one function, no analytic inverse
exists.  Hence, we seek an image $x$ that, when we compute HOG on it, closely
matches the original descriptor $y$: \begin{align} \phi^{-1}(y) = \argmin_{x
\in \mathbb{R}^{D}} \left|\left| \phi(x) - y \right|\right|_2^2
\label{eqn:objective} \end{align} Optimizing Eqn.\ref{eqn:objective} is
challenging. Although Eqn.\ref{eqn:objective} is not convex, we tried
gradient-descent strategies by numerically evaluating the derivative in image
space with Newton's method.  Unfortunately, we observed poor results, likely
because HOG is both highly sensitive to noise and Eqn.\ref{eqn:objective} has
frequent local minima.

In the rest of this section, we present four algorithms for inverting HOG
features. Since, to our knowledge, no algorithms to invert HOG have yet been
developed, we first describe three simple baselines for HOG inversion. In the
last section, we present our main inversion algorithm.


\subsection{Baseline A: Exemplar LDA (ELDA)}

\begin{figure}
\centering
\includegraphics[width=\linewidth]{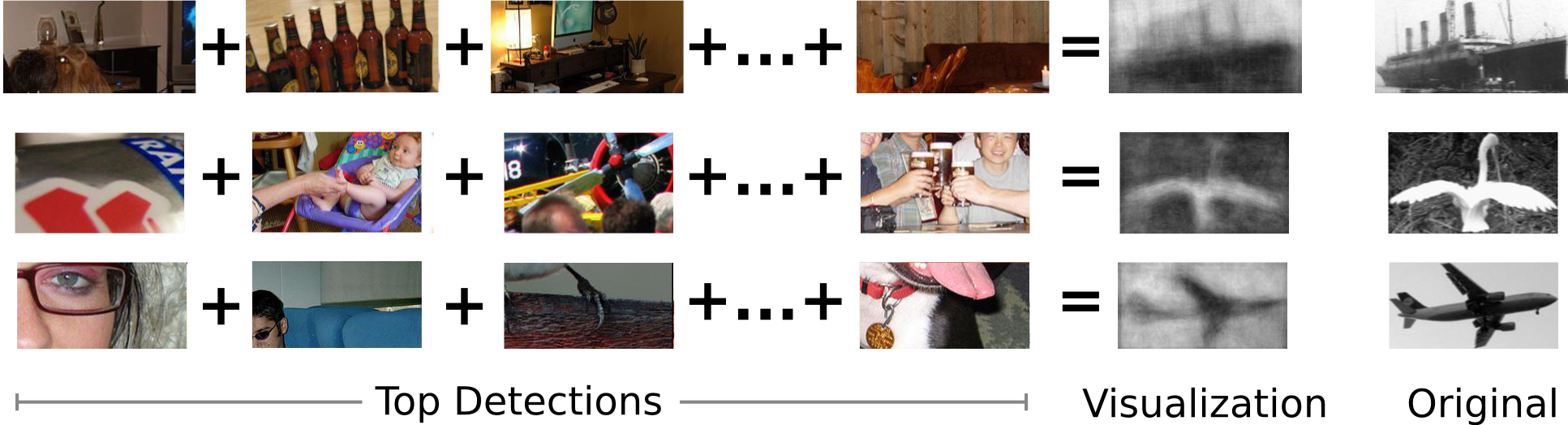}

\caption{We found that averaging the images of top detections from an exemplar
LDA detector provide one method to invert HOG features.}

\label{fig:elda}

\vspace{-1em}

\end{figure}

Consider the top detections for the exemplar object detector
\cite{hariharan2012discriminative, malisiewicz2011ensemble} for a few
images shown in Figure \ref{fig:elda}. Although all top detections are false
positives, notice that each detection captures some statistics about the query.
Even though the detections are wrong, if we squint, we can see parts of the
original object appear in each detection.

We use this simple observation to produce our first inversion baseline.
Suppose we wish to invert HOG feature $y$. We first train an exemplar LDA
detector \cite{hariharan2012discriminative} for this query, $w = \Sigma^{-1}(y
- \mu)$. We score $w$ against every sliding window on a large database. The HOG
inverse is then the average of the top $K$ detections in RGB space:
$\phi^{-1}_{A}(y) = \frac{1}{K} \sum_{i=1}^K z_i$ where $z_i$ is an image of a
top detection.

This method, although simple, produces surprisingly accurate reconstructions,
even when the database does not contain the category of the HOG template.
However, it is computationally expensive since it requires running an object detector across a large database.
We also
point out that a similar nearest neighbor method is used in brain research to
visualize what a person might be seeing \cite{nishimoto2011reconstructing}.

\subsection{Baseline B: Ridge Regression}

We present a fast, parametric
inversion baseline based off ridge regression.
Let $X \in \mathbb{R}^D$ be a random variable representing a gray scale image and $Y \in \mathbb{R}^d$ be a random variable of its corresponding HOG point. We define
these random variables to be normally distributed on a $D+d$-variate Gaussian
$P(X, Y) \sim \mathcal{N}(\mu, \Sigma)$
with parameters
$\mu = \left[\begin{smallmatrix}
\mu_X & \mu_{Y}
\end{smallmatrix}\right]$ and $
\Sigma = \left[\begin{smallmatrix}
\Sigma_{XX} & \Sigma_{XY} \\
\Sigma_{XY}^T & \Sigma_{YY}
\end{smallmatrix}\right]$.
In order to invert a HOG feature $y$, we calculate the most likely image from
the conditional Gaussian distribution $P(X | Y = y)$:
\begin{align}
\phi^{-1}_{B}(y) &= \argmax_{x \in \mathbb{R}^D} P(X = x | Y = y)
\end{align}
It is well known that Gaussians have a closed form conditional mode: 
\begin{align}
\phi^{-1}_{B}(y) = \Sigma_{XY} \Sigma_{YY}^{-1} (y - \mu_{Y}) + \mu_X
\end{align}
Under this inversion algorithm, any HOG point can be inverted by a single matrix
multiplication, allowing for inversion in under a second. 

We estimate $\mu$ and $\Sigma$ on a large database. In practice, $\Sigma$ is not positive definite; we add
a small uniform prior (i.e., $\hat{\Sigma} = \Sigma + \lambda I$) so $\Sigma$
can be inverted. Since we wish to invert any HOG point, we assume that $P(X,
Y)$ is stationary \cite{hariharan2012discriminative}, allowing us to efficiently learn the covariance across
massive datasets. We invert an arbitrary dimensional HOG point by marginalizing
out unused dimensions.

We found that ridge regression yields blurred inversions.  Intuitively, since
HOG is invariant to shifts up to its bin size, there are many images that
map to the same HOG point. Ridge regression is reporting the statistically most
likely image, which is the average over all shifts. This causes ridge
regression to only recover the low frequencies of the original image.

\subsection{Baseline C: Direct Optimization}

We now provide a baseline that attempts to find images that, when we compute HOG on it, sufficiently match the original descriptor.  In
order to do this efficiently, we only consider images that span a natural image basis.
Let $U \in \mathbb{R}^{D \times K}$ be the natural image basis. We found using the first $K$ eigenvectors of $\Sigma_{XX} \in \mathbb{R}^{D \times D}$ worked well for this basis. Any image $x \in \mathbb{R}^D$ can be encoded by coefficients $\rho \in \mathbb{R}^K$ in this basis: $x = U \rho$. We wish to minimize:
\begin{equation}
\begin{aligned}
&\phi^{-1}_{C}(y) = U\rho^* \\
&\textrm{where} \quad \rho^* = \argmin_{\rho \in \mathbb{R}^K} \left|\left| \phi(U\rho) - y \right|\right|_2^2
\end{aligned}
\label{eqn:highfreq-objective}
\end{equation}
Empirically we found success optimizing Eqn.\ref{eqn:highfreq-objective} using
coordinate descent on $\rho$ with random restarts.
We use an over-complete basis corresponding to sparse Gabor-like filters for
$U$. We compute the eigenvectors of $\Sigma_{XX}$ across different scales and
translate smaller eigenvectors to form $U$.

\subsection{Algorithm D: Paired Dictionary Learning}

\begin{figure}
\includegraphics[width=\linewidth]{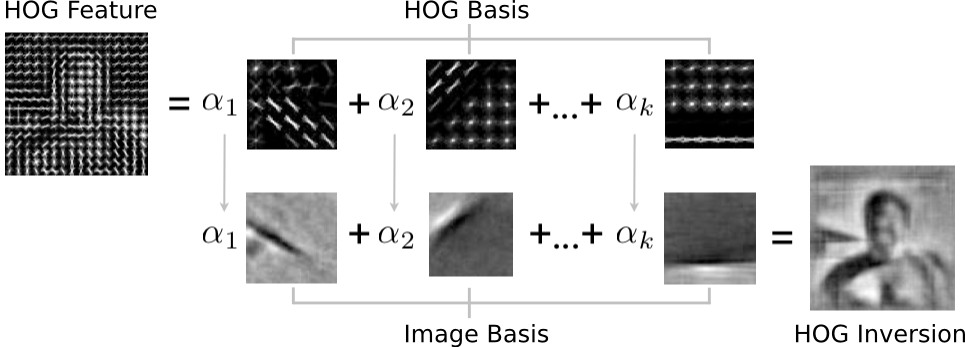}
\caption{Inverting HOG using paired dictionary learning. We first project the HOG vector
on to a HOG basis. By jointly learning a coupled
basis of HOG features and natural images, we then transfer the coefficients
to the image basis to recover the natural image.}
\label{fig:pair-tutorial}
\end{figure}

\begin{figure}
\includegraphics[width=1\linewidth]{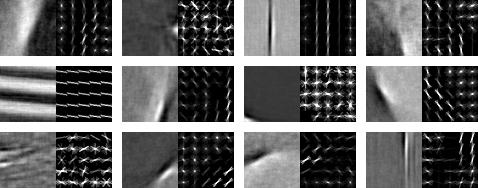}
\caption{Some pairs of dictionaries for $U$ and $V$. The left of every pair is the gray scale dictionary element and the right is the positive components elements in the HOG dictionary. Notice the correlation between dictionaries.}
\label{fig:pair-basis}
\vspace{-1em}
\end{figure}

In this section, we present our main inversion algorithm. 
Let $x \in \mathbb{R}^{D}$ be an image and $y \in \mathbb{R}^{d}$ be its HOG descriptor. Suppose we write $x$ and $y$ in terms  of bases $U \in \mathbb{R}^{D \times K}$ and $V \in \mathbb{R}^{d \times K}$ respectively, but with shared coefficients $\alpha \in \mathbb{R}^K$:
\begin{align}
x = U \alpha \quad \textrm{and} \quad y = V \alpha \label{eqn:pair}
\end{align}
The key observation is that inversion can be obtained by first projecting the HOG features $y$ onto the HOG basis $V$, then projecting $\alpha$ into the natural image basis $U$:
\begin{equation}
\begin{aligned}
&\phi^{-1}_{D}(y) = U\alpha^* \\
&\textrm{where} \quad \alpha^* = \argmin_{\alpha \in \mathbb{R}^K} ||V\alpha-y||_2^2 \quad \textrm{s.t.} \quad ||\alpha||_1 \le \lambda
\end{aligned}
\label{eqn:pair-inverse}
\end{equation}
See Figure \ref{fig:pair-tutorial} for a graphical representation of the paired dictionaries.
Since efficient solvers for Eqn.\ref{eqn:pair-inverse} exist \cite{mairal2009online,lee2007efficient}, we can invert features in under two seconds on a 4 core CPU.

Paired dictionaries require finding appropriate bases $U$ and $V$ such that Eqn.\ref{eqn:pair} holds. To do this, we solve a paired dictionary learning problem, inspired by recent super resolution sparse coding work \cite{yang2010image,wang2012semi}:
\begin{equation}
\begin{split}
\argmin_{U, V, \alpha} \; & \sum_{i=1}^N \left(||x_i - U\alpha_i||_2^2 + ||\phi(x_i) - V\alpha_i||_2^2\right) \\
\textrm{s.t.} \quad &||\alpha_i||_1 \le \lambda \; \forall i, \; ||U||_2^2 \le \gamma_1, \; ||V||_2^2 \le \gamma_2 \
\end{split}
\label{eqn:pairobj}
\end{equation}
After a few algebraic manipulations, the above objective simplifies to a
standard sparse coding and dictionary learning problem with concatenated
dictionaries, which we optimize using SPAMS \cite{mairal2009online}.
Optimization typically took a few hours on medium sized problems. We estimate
$U$ and $V$ with a dictionary size $K \approx 10^3$ and training samples $N \approx 10^6$ from a large database. See Figure \ref{fig:pair-basis} for a visualization of
the learned dictionary pairs.

Unfortunately, the paired dictionary learning formulation suffers on problems
of nontrivial scale. In practice, we only learn
dictionaries for $5 \times 5$ HOG templates. In order to
invert a $w \times h$ HOG template $y$, we invert every $5 \times 5$ subpatch
inside $y$ and average overlapping patches in the final reconstruction. We
found that this approximation works well in practice.

\section{Evaluation of Visualizations}

We evaluate our inversion algorithms using both qualitative
and quantitative measures. We use PASCAL VOC 2011 \cite{Everingham10} as our
dataset and we invert patches corresponding to objects. Any algorithm that
required training could only access the training set. During evaluation, only
images from the validation set are examined.  The database for exemplar LDA
excluded the category of the patch we were inverting to reduce the potential effect of
dataset biases.

\begin{figure}
\vspace{-1em}
\centering \subfloat[Original]{ \shortstack{
\includegraphics[width=0.18\linewidth]{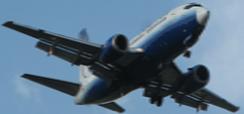} \\
\includegraphics[width=0.18\linewidth]{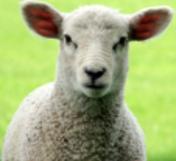} \\
\includegraphics[width=0.18\linewidth]{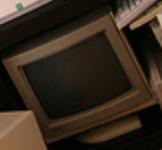}\\
\includegraphics[width=0.18\linewidth]{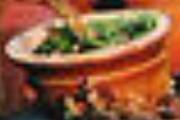} \\
\includegraphics[width=0.18\linewidth]{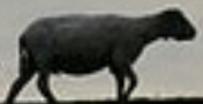}\\
\includegraphics[width=0.18\linewidth]{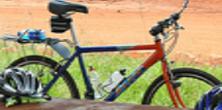} \\
\includegraphics[width=0.18\linewidth]{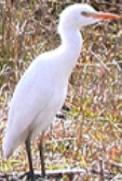} \\
\includegraphics[width=0.18\linewidth]{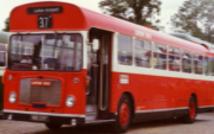} \\
\includegraphics[width=0.18\linewidth]{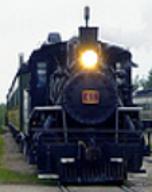}\\
\includegraphics[width=0.18\linewidth]{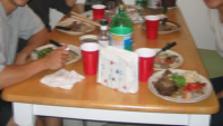}
} }
\subfloat[ELDA]{ \shortstack{
\includegraphics[width=0.18\linewidth]{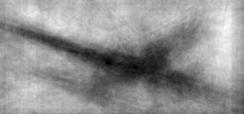} \\
\includegraphics[width=0.18\linewidth]{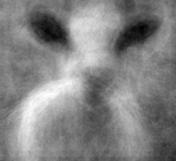} \\
\includegraphics[width=0.18\linewidth]{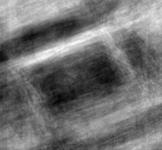} \\
\includegraphics[width=0.18\linewidth]{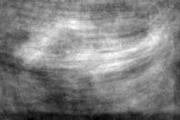} \\
\includegraphics[width=0.18\linewidth]{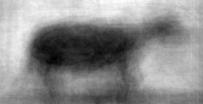} \\
\includegraphics[width=0.18\linewidth]{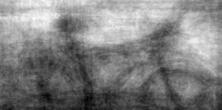} \\
\includegraphics[width=0.18\linewidth]{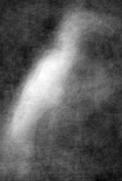} \\
\includegraphics[width=0.18\linewidth]{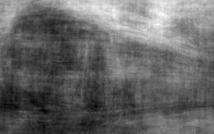} \\
\includegraphics[width=0.18\linewidth]{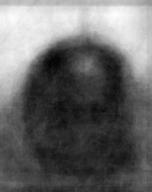}\\
\includegraphics[width=0.18\linewidth]{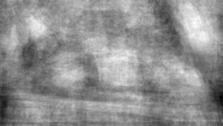}
} }
\subfloat[Ridge]{ \shortstack{
\includegraphics[width=0.18\linewidth]{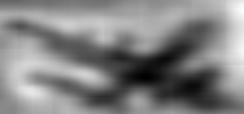} \\
\includegraphics[width=0.18\linewidth]{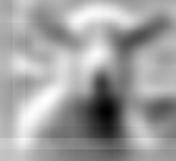}\\
\includegraphics[width=0.18\linewidth]{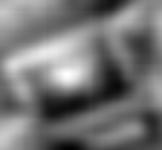} \\
\includegraphics[width=0.18\linewidth]{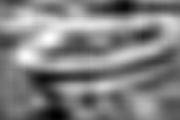} \\
\includegraphics[width=0.18\linewidth]{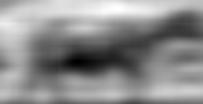} \\
\includegraphics[width=0.18\linewidth]{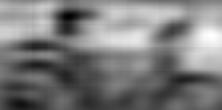} \\
\includegraphics[width=0.18\linewidth]{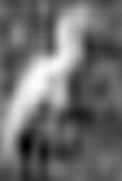} \\
\includegraphics[width=0.18\linewidth]{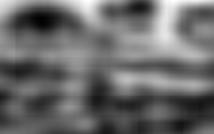}\\
\includegraphics[width=0.18\linewidth]{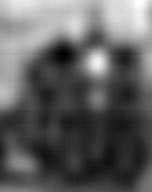}\\
\includegraphics[width=0.18\linewidth]{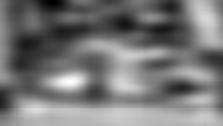}
} }
\subfloat[Direct]{ \shortstack{
\includegraphics[width=0.18\linewidth]{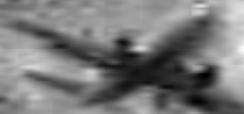} \\
\includegraphics[width=0.18\linewidth]{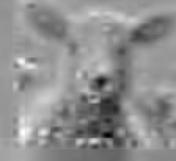} \\
\includegraphics[width=0.18\linewidth]{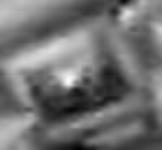} \\
\includegraphics[width=0.18\linewidth]{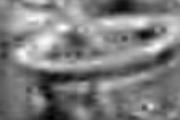}\\
\includegraphics[width=0.18\linewidth]{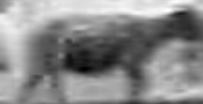}\\
\includegraphics[width=0.18\linewidth]{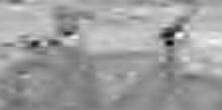}\\
\includegraphics[width=0.18\linewidth]{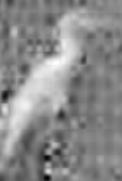} \\
\includegraphics[width=0.18\linewidth]{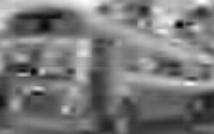}\\
\includegraphics[width=0.18\linewidth]{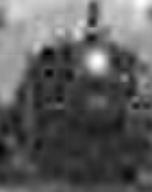}\\
\includegraphics[width=0.18\linewidth]{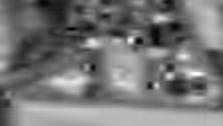}
} }
\subfloat[PairDict]{ \shortstack{
\includegraphics[width=0.18\linewidth]{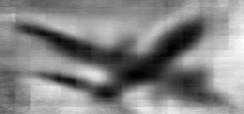} \\
\includegraphics[width=0.18\linewidth]{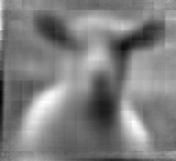} \\
\includegraphics[width=0.18\linewidth]{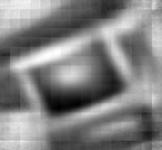} \\
\includegraphics[width=0.18\linewidth]{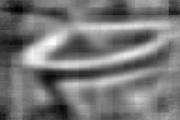}\\
\includegraphics[width=0.18\linewidth]{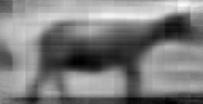}\\
\includegraphics[width=0.18\linewidth]{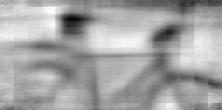}\\
\includegraphics[width=0.18\linewidth]{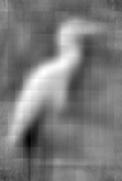} \\
\includegraphics[width=0.18\linewidth]{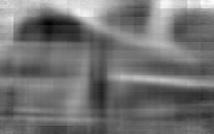}\\
\includegraphics[width=0.18\linewidth]{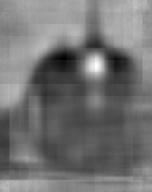}\\
\includegraphics[width=0.18\linewidth]{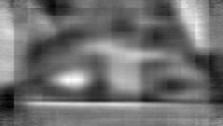}
} } \caption{We
show the results for all four of our inversion algorithms on held out image
patches on similar dimensions common for object detection.}
\label{fig:results}

\vspace{-0.5em}
\end{figure}

We show our inversions in Figure \ref{fig:results} for a few object categories.
Exemplar LDA and ridge regression tend to produce blurred visualizations.
Direct optimization recovers high frequency details at the expense of extra
noise. Paired dictionary learning tends to produce the best visualization
for HOG descriptors. By learning a dictionary over the visual world and
the correlation between HOG and natural images, paired dictionary learning
recovered high frequencies without introducing significant noise.



\begin{figure}

\centering

\includegraphics[width=\linewidth]{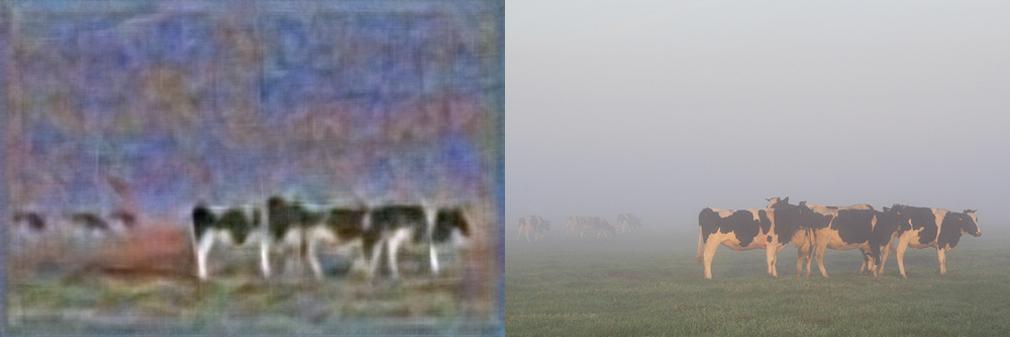}

\caption{We show results where our paired dictionary algorithm is trained to
recover RGB images instead of only grayscale images. The right shows the
original image and the left shows the inverse.}

\label{fig:color}
\vspace{-1em}

\end{figure}

We discovered that the paired dictionary is able to recover color from HOG
descriptors.  Figure \ref{fig:color} shows the result of training a paired
dictionary to estimate RGB images instead of grayscale images. While the paired
dictionary assigns arbitrary colors to man-made objects and in-door scenes, it
frequently colors natural objects correctly, such as grass or the sky, likely
because those categories are strongly correlated to HOG descriptors. We focus
on grayscale visualizations in this paper because we found those to be more
intuitive for humans to understand.

\begin{figure}
\centering
\includegraphics[width=0.25\linewidth]{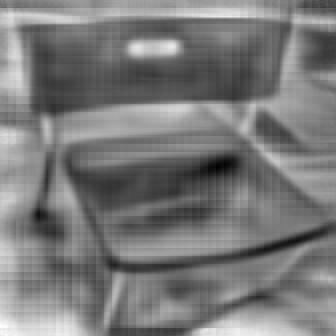}\includegraphics[width=0.25\linewidth]{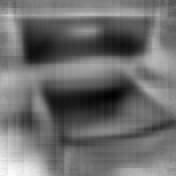}\includegraphics[width=0.25\linewidth]{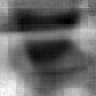}\includegraphics[width=0.25\linewidth]{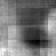}

\caption{Our inversion algorithms are sensitive to the HOG template size.
Larger templates are easier to invert since they are less invariant. We show
how performance degrades as the template becomes smaller. Dimensions in HOG
space shown: $40 \times 40$, $20 \times 20$, $10 \times 10$, and $5 \times 5$.}

\label{fig:pyramid}

\vspace{-1.5em}

\end{figure}

Our HOG inversions are sensitive to the dimension of their
templates. For medium ($10 \times 10$) to large templates ($40 \times 40$), we
obtain reasonable performance.  But, for small templates ($5 \times 5$) the
inversion is blurred.  Fig.\ref{fig:pyramid} shows examples as the HOG
descriptor dimension changes.

In the remainder of this section, we evaluate our algorithms under two
benchmarks. Firstly, we evaluate against an automatic inversion metric that
measures how well our inversions reconstruct the original images. Secondly, we
conducted a large visualization challenge with human subjects on Amazon
Mechanical Turk (MTurk), which is designed to determine how well people can infer the original
category from the inverse.  The first experiment measures the algorithm's
reconstruction error, while the second experiment analyzes the people's ability
to recover high level semantics from our visualizations.

\subsection{Inversion Benchmark}

We consider the inversion performance of our algorithm: given a HOG feature
$y$, how well does our inverse $\phi^{-1}(y)$ reconstruct the original pixels
$x$ for each algorithm? Since HOG is invariant up to a constant shift and
scale, we score each inversion against the original image with normalized cross
correlation. Our results are shown in Table \ref{tab:inversionobjective}.
Overall, exemplar LDA does the best at pixel level reconstruction.

\begin{table}
\centering
\begin{tabular}{l | c c c c c}
Category & ELDA & Ridge & Direct & PairDict \\
\hline
bicycle &        0.452 &  \textbf{0.577} &  0.513 &  0.561 \\
bottle &         \textbf{0.697} &  0.683 &  0.660 &  0.671 \\
car &    0.668 &  \textbf{0.677} &  0.652 &  0.639 \\
cat &    \textbf{0.749} &  0.712 &  0.687 &  0.705 \\
chair &  \textbf{0.660} &  0.621 &  0.604 &  0.617 \\
table &    \textbf{0.656} &  0.617 &  0.582 &  0.614 \\
horse &  \textbf{0.686} &  0.633 &  0.586 &  0.635 \\
motorbike &      0.573 &  \textbf{0.617} &  0.549 &  0.592 \\
person &         \textbf{0.696} &  0.667 &  0.646 &  0.646 \\
\hline
Mean & \textbf{0.671} & 0.656  &0.620 & 0.637\\
\end{tabular}

\caption{We evaluate the performance of our inversion algorithm by comparing
the inverse to the ground truth image using the mean normalized cross
correlation. Higher is better; a score of 1 is perfect.}

\label{tab:inversionobjective}

\vspace{-0.5em}

\end{table}

\setlength{\tabcolsep}{2pt}
\begin{table}
\centering
\begin{tabular}{l | c c c c c | c}
Category & ELDA & Ridge & Direct & PairDict & Glyph & Expert \\
\hline
bicycle & 0.327& 0.127& 0.362& 0.307& \textbf{0.405} & 0.438 \\
bottle & 0.269& 0.282& 0.283& \textbf{0.446}& 0.312 & 0.222\\
car & 0.397& 0.457& \textbf{0.617}& 0.585& 0.359 & 0.389\\
cat & 0.219& 0.178& \textbf{0.381}& 0.199& 0.139 & 0.286 \\
chair & 0.099& 0.239& 0.223& \textbf{0.386}& 0.119 & 0.167\\
table & 0.152& 0.064& 0.162& \textbf{0.237}& 0.071 & 0.125\\
horse & 0.260& 0.290& 0.354& \textbf{0.446}& 0.144 & 0.150\\
motorbike & 0.221& 0.232& \textbf{0.396}& 0.224& 0.298 & 0.350\\
person & 0.458& 0.546& 0.502& \textbf{0.676}& 0.301 & 0.375\\
\hline
Mean & 0.282& 0.258& 0.355& \textbf{0.383} & 0.191 & 0.233 
\end{tabular}

\caption{We evaluate visualization performance across twenty PASCAL VOC
categories by asking MTurk workers to classify our inversions.  Numbers are
percent classified correctly; higher is better. Chance is $0.05$.  Glyph refers
to the standard black-and-white HOG diagram popularized by
\cite{dalal2005histograms}.  Paired dictionary learning provides the best
visualizations for humans. Expert refers to PhD students in computer vision
performing the same visualization challenge with HOG glyphs.}

\label{tab:userstudy}
\vspace{-1em}
\end{table}
%


\subsection{Visualization Benchmark}

While the inversion benchmark evaluates how well the inversions reconstruct the
original image, it does not capture the high level content of the inverse: is
the inverse of a sheep still a sheep? To evaluate this, we conducted a study on
MTurk. We sampled 2,000 windows corresponding to objects in
PASCAL VOC 2011. We then showed participants an inversion from one of our
algorithms and asked users to classify it into one of the 20 categories. Each
window was shown to three different users. Users were required to pass a
training course and qualification exam before participating in order to
guarantee users understood the task. Users could optionally select that they
were not confident in their answer. We also compared our algorithms against the
standard black-and-white HOG glyph popularized by \cite{dalal2005histograms}.

Our results in Table \ref{tab:userstudy} show that paired dictionary learning
and direct optimization provide the best visualization of HOG descriptors for
humans.  Ridge regression and exemplar LDA performs better than the glyph, but
they suffer from blurred inversions.  Human performance on the HOG glyph is
generally poor, and participants were even the slowest at completing that
study. Interestingly, the glyph does the best job at visualizing bicycles,
likely due to their unique circular gradients. Our results overall suggest that
visualizing HOG with the glyph is misleading, and richer visualizations from
our paired dictionary are useful for interpreting HOG vectors.



There is strong correlation with the accuracy of humans classifying the HOG
inversions with the performance of HOG based object detectors.  We found human
classification accuracy on inversions and the state-of-the-art object detection
AP scores from \cite{felzenszwalb2010cascade} are correlated with a Spearman's
rank correlation coefficient of 0.77. This result suggests that humans can
predict the performance of object detectors by only looking at HOG
visualizations.

We also asked computer vision PhD students to classify HOG glyphs in order to
compare MTurk workers with experts in HOG. Our results are summarized
in the last column of Table \ref{tab:userstudy}.  HOG experts performed slightly
better than non-experts on the glyph challenge, but experts on glyphs did not
beat non-experts on other visualizations. This result suggests that our
algorithms produce more intuitive visualizations even for object detection
researchers.

\section{Understanding Object Detectors}

\begin{figure}
\centering
\vspace{-1em}
\subfloat[Human Vision]{
\includegraphics[height=16em]{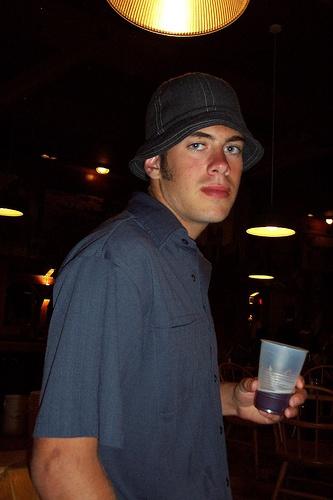}
\label{fig:seeintodarkA}
}
\subfloat[HOG Vision]{
\includegraphics[height=16em]{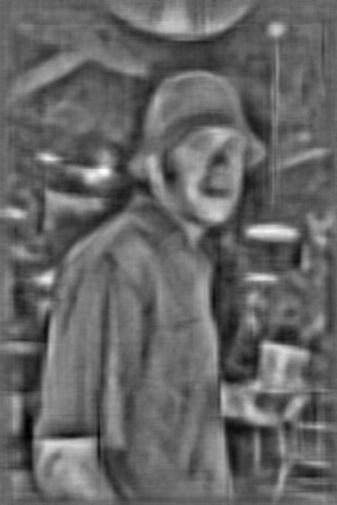}
\label{fig:seeintodarkB}
}

\caption{HOG inversion reveals the world that object detectors see. The left
shows a man standing in a dark room. If we compute HOG on this
image and invert it, the previously dark scene behind the man emerges. Notice
the wall structure, the lamp post, and the chair in the bottom right hand
corner.}

\label{fig:seeintodark}

\vspace{-1em}

\end{figure}

We have so far presented four algorithms to visualize object detection
features. We evaluated the visualizations with a large human study, and we
found that paired dictionary learning provides the most intuitive visualization
of HOG features.  In this section, we will use this visualization to inspect
the behavior of object detection systems.

\subsection{HOG Goggles}

Our visualizations reveal that the world that features see is slightly
different from the world that the human eye perceives. Figure \ref{fig:seeintodarkA} shows a normal
photograph of a man standing in a dark room, but Figure \ref{fig:seeintodarkB}
shows how HOG features see the same man. Since HOG is invariant to illumination
changes and amplifies gradients, the background of the scene, normally
invisible to the human eye, materializes in our visualization. 

In order to understand how this clutter affects object detection, we visualized
the features of some of the top false alarms from the Felzenszwalb et al.\
object detection system \cite{felzenszwalb2010object} when applied to the
PASCAL VOC 2007 test set.  Figure \ref{fig:topdets} shows our visualizations of
the features of the top false alarms. Notice how the false alarms look very
similar to true positives. This result suggests that these failures are due to
limitations of HOG, and consequently, even if we develop better learning
algorithms or use larger datasets, these will false alarms will likely persist.

\begin{figure*}
\centering

\includegraphics[width=\linewidth]{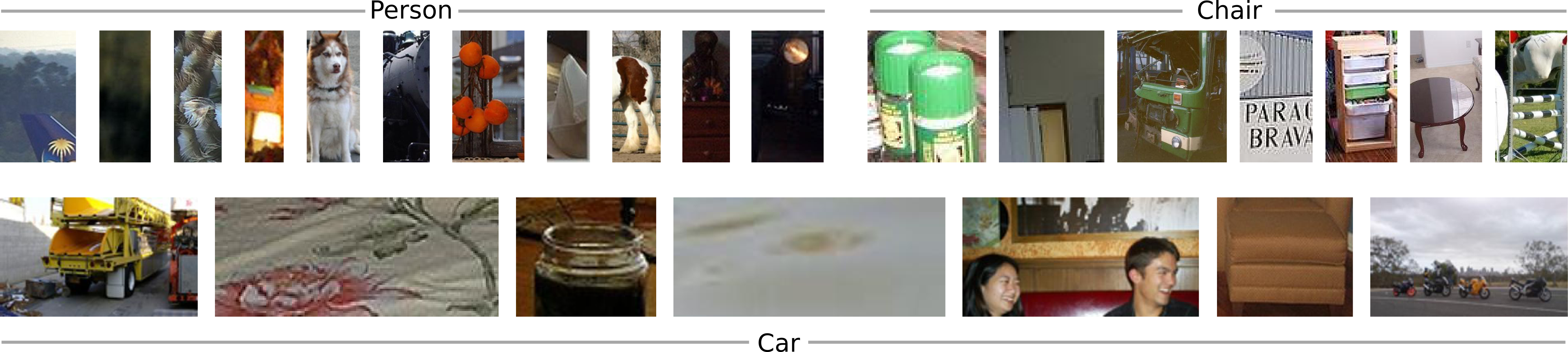}

\caption{We show the original RGB patches that correspond to the
visualizations from Figure \ref{fig:topdets}.  We print the original patches on
a separate page to highlight how the inverses of false positives look
like true positives. We recommend comparing this figure side-by-side with
Figure \ref{fig:topdets}.  Top left: person detections, top right: chair
detections, bottom: car detections.} \label{fig:topdetsrgb}

\vspace{-1.5em}

\end{figure*}

Figure \ref{fig:topdetsrgb} shows the corresponding RGB image patches for the
false positives discussed above. Notice how when we view these detections in
image space, all of the false alarms are difficult to explain. Why do chair
detectors fire on buses, or people detectors on cherries? Instead, by
visualizing the detections in feature space, we discovered that the learning
algorithm actually made reasonable failures since the features are deceptively
similar to true positives.

\subsection{Human+HOG Detectors} 

Although HOG features are designed for machines, how well do humans see in HOG
space? If we could quantify human vision on the HOG feature space, we could get
insights into the performance of HOG with a perfect learning algorithm
(people). Inspired by Parikh and Zitnick's methodology
\cite{parikh2011human,parikh2010role}, we conducted a large human study where
we had Amazon Mechanical Turk workers act as sliding window HOG based object
detectors.

We built an online interface for humans to look at HOG visualizations of window
patches at the same resolution as DPM. We instructed workers to either classify
a HOG visualization as a positive example or a negative example for a category.
By averaging over multiple people (we used 25 people per window), we obtain a
real value score for a HOG patch. To build our dataset, we sampled top
detections\footnote{Note that recall will now go to $1$ in these experiments
because we only consider windows that DPM detects. Consequently, this
experiment can only give us relative orderings of detectors. Unfortunately,
computing full precision-recall curves is cost prohibitive.} from DPM on the
PASCAL VOC 2007 dataset for a few categories. Our dataset consisted of around $5,000$
windows per category and around $20\%$ were true positives.

\begin{figure}
\includegraphics[width=0.5\linewidth]{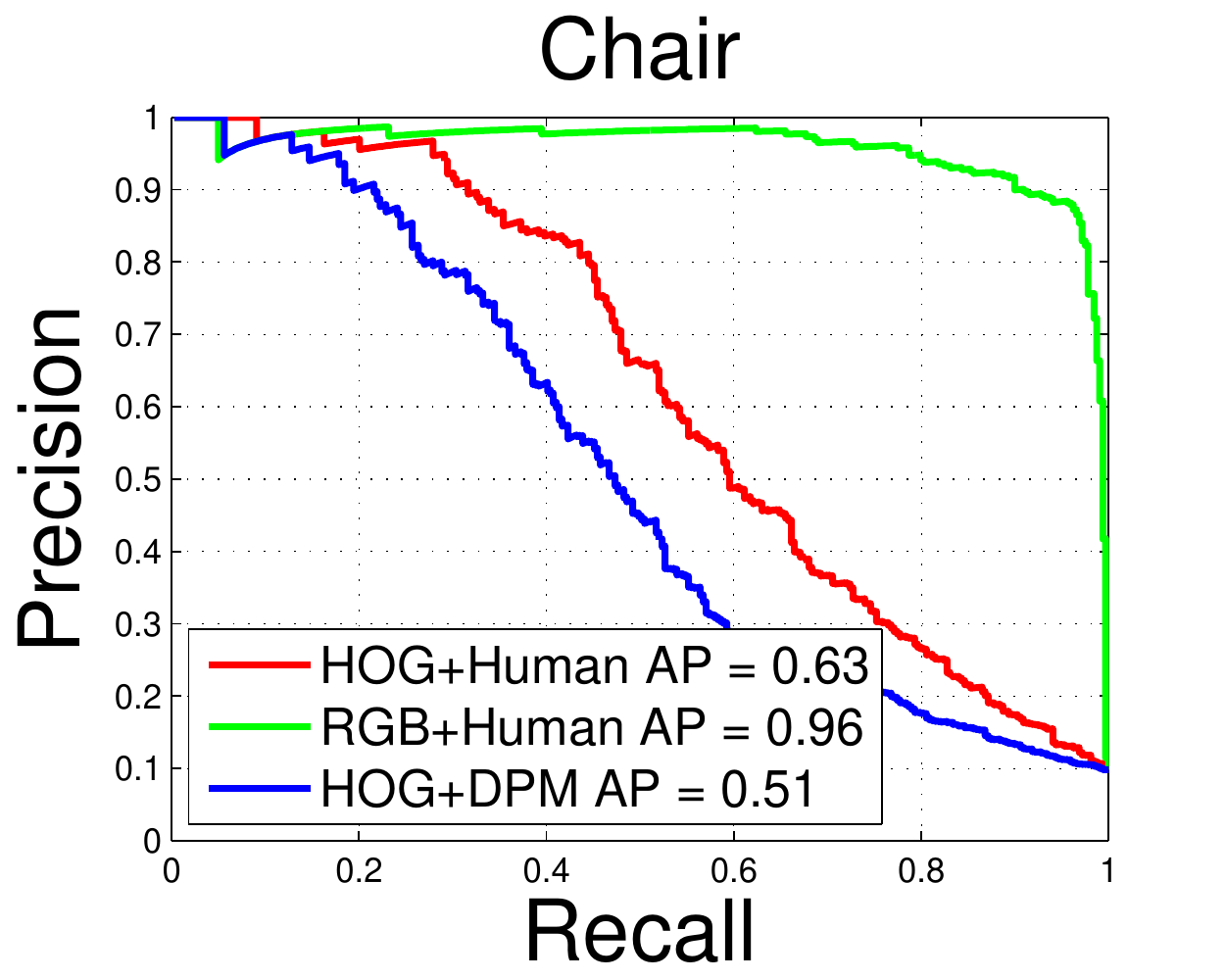}\includegraphics[width=0.5\linewidth]{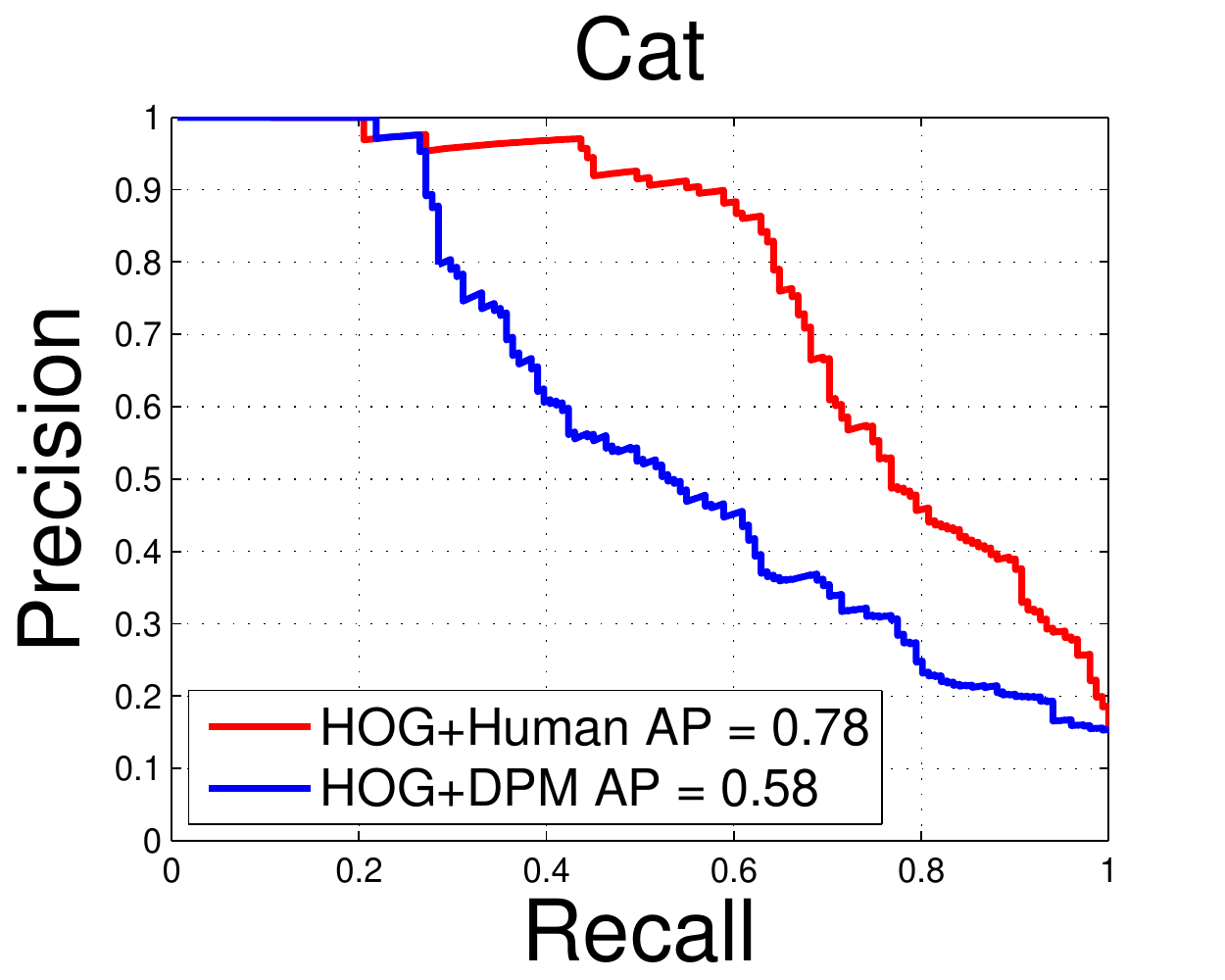}

\includegraphics[width=0.5\linewidth]{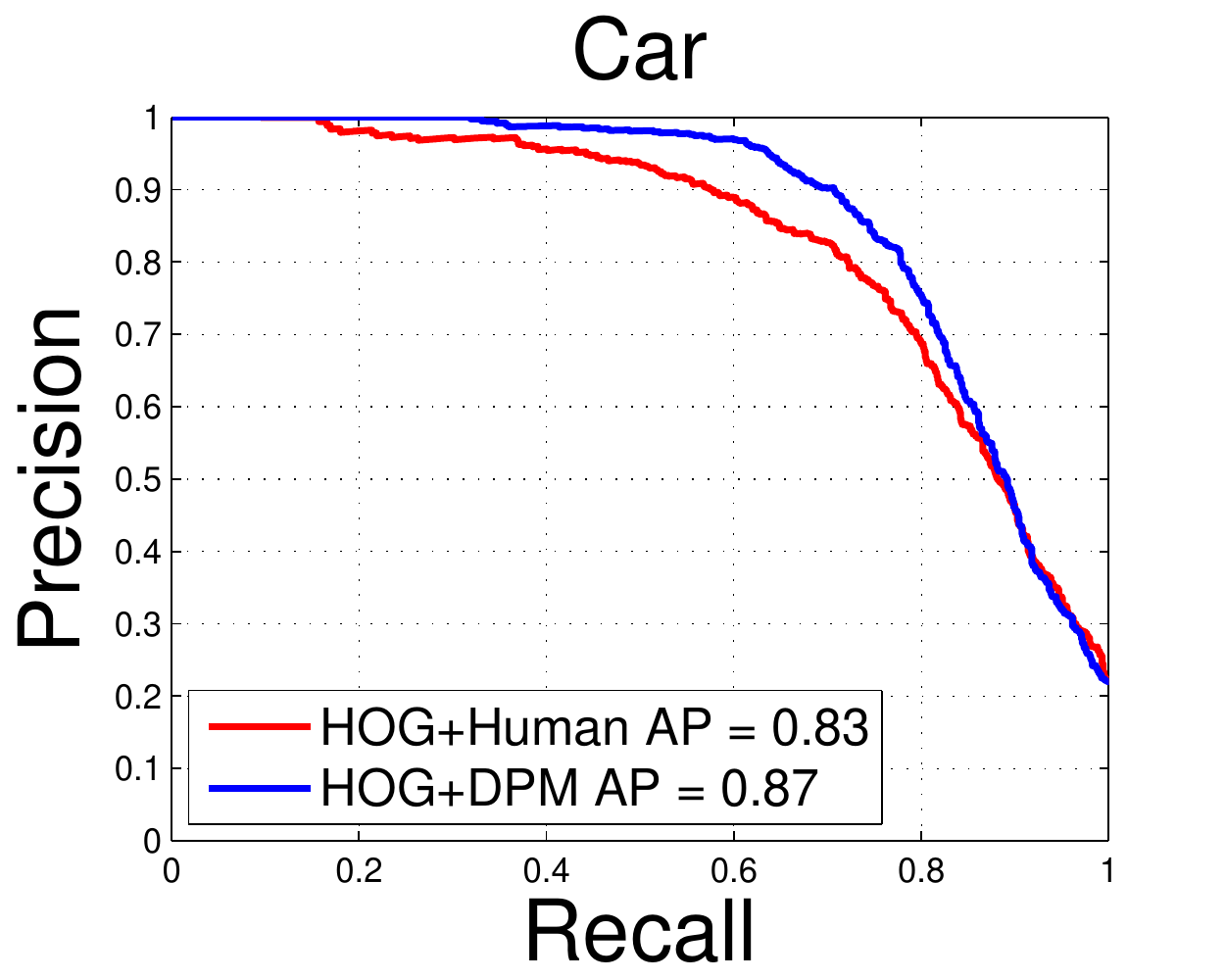}\includegraphics[width=0.5\linewidth]{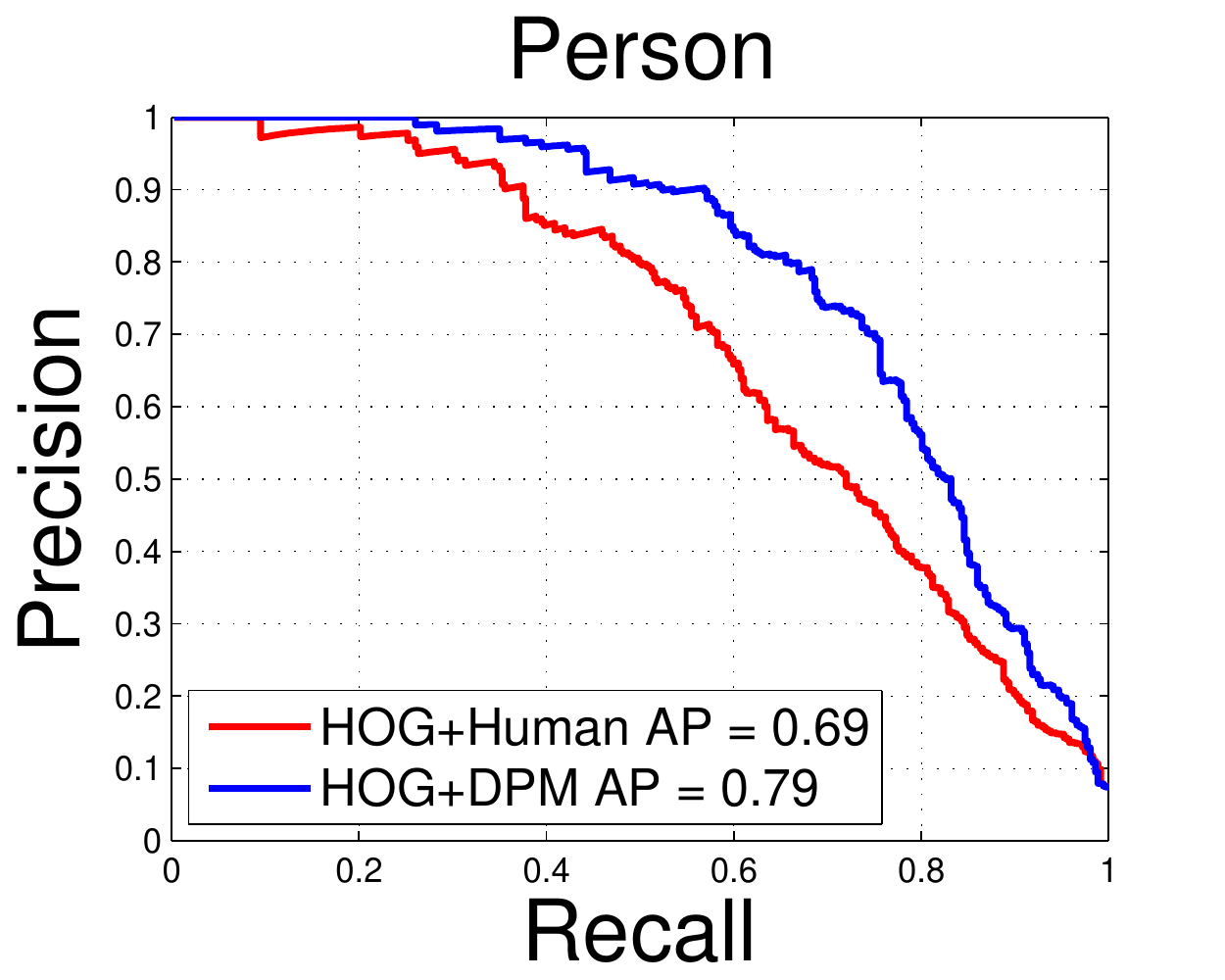}

\caption{By instructing multiple human subjects to classify the visualizations, we show
performance results with an ideal learning algorithm (i.e., humans) on the HOG
feature space. Please see text for details.}

\label{fig:hoggles}

\vspace{-1.5em}

\end{figure}

Figure \ref{fig:hoggles} shows precision recall curves for the Human+HOG based
object detector. In most cases, human subjects classifying HOG visualizations
were able to rank sliding windows with either the same accuracy or better than
DPM. Humans tied DPM for recognizing cars, suggesting that performance may be
saturated for car detection on HOG.  Humans were slightly superior to DPM for
chairs, although performance might be nearing saturation soon. There appears to
be the most potential for improvement for detecting cats with HOG. Subjects
performed slightly worst than DPM for detecting people, but we believe this is
the case because humans tend to be good at fabricating people in abstract
drawings.

We then repeated the same experiment as above on chairs except we instructed
users to classify the original RGB patch instead of the HOG visualization. As
expected, humans achieved near perfect accuracy at detecting chairs with
RGB sliding windows. The performance gap between the Human+HOG detector and Human+RGB detector
demonstrates the amount of information that HOG features discard. 

%
%
%
%

Our experiments suggest that there is still some performance left to be
squeezed out of HOG. However, DPM is likely operating very close to the
performance limit of HOG. Since humans are the ideal learning agent and they
still had trouble detecting objects in HOG space, HOG may be too lossy of a
descriptor for high performance object detection.  If we wish to significantly
advance the state-of-the-art in recognition, we suspect focusing effort on
building better features that capture finer details as well as higher level
information will lead to substantial performance improvements in object
detection.

\subsection{Model Visualization}

\begin{figure*}
\centering
\includegraphics[height=5.7em]{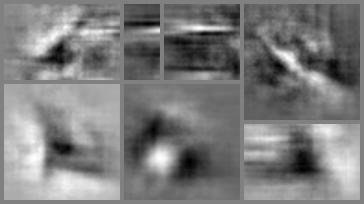}\hspace{0.5em}
\includegraphics[height=5.7em]{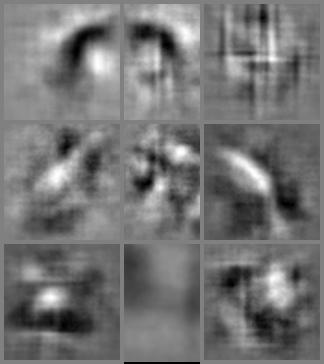}\hspace{0.5em}
\includegraphics[height=5.7em]{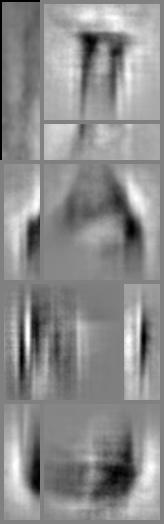}\hspace{0.5em}
\includegraphics[height=5.7em]{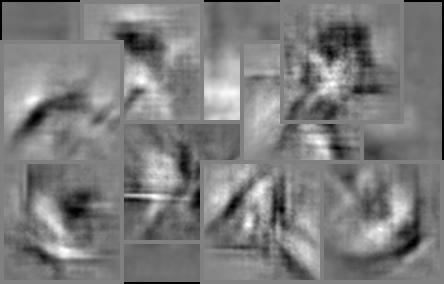}\hspace{0.5em}
\includegraphics[height=5.7em]{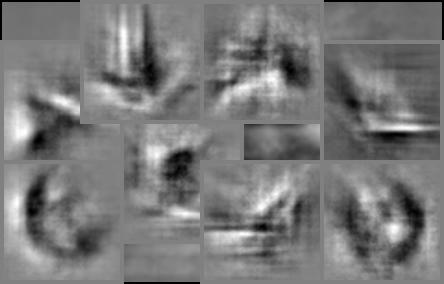}\hspace{0.5em}
\includegraphics[height=5.7em]{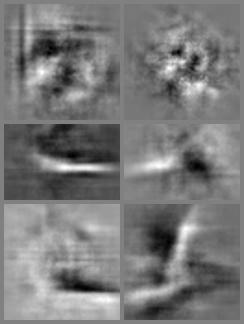}\hspace{0.1em}\includegraphics[height=5.7em]{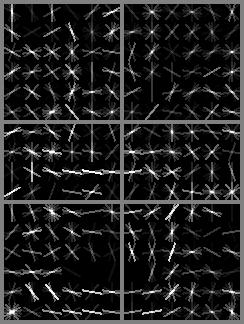} \\
\vspace{0.5em}
\includegraphics[height=5.7em]{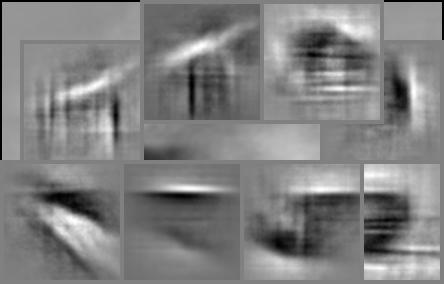}\hspace{0.5em}
\includegraphics[height=5.7em]{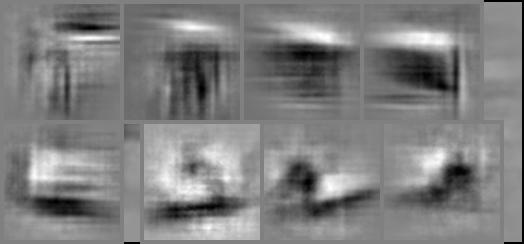}\hspace{0.5em}
\includegraphics[height=5.7em]{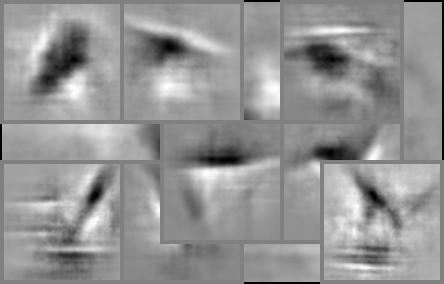}\hspace{0.5em}
\includegraphics[height=5.7em]{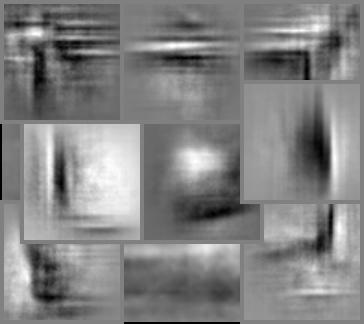}\hspace{0.5em}
\includegraphics[height=5.7em]{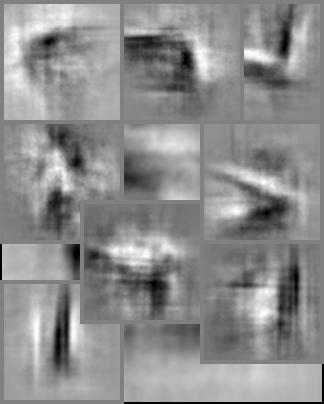}\hspace{0.1em}\includegraphics[height=5.7em]{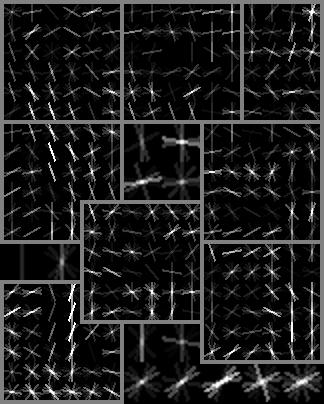}

\caption{We visualize a few deformable parts models trained with
\cite{felzenszwalb2010object}. Notice the structure that emerges with our
visualization.  First row: car, person, bottle, bicycle, motorbike, potted
plant. Second row: train, bus, horse, television, chair. For the right most
visualizations, we also included the HOG glyph. Our visualizations
tend to reveal more detail than the glyph.}

\label{fig:prototypes}

\vspace{-0.5em}

\end{figure*}

We found our algorithms are also useful for visualizing the learned models of an
object detector.  Figure \ref{fig:prototypes} visualizes the root templates and
the parts from \cite{felzenszwalb2010object} by inverting the
positive components of the learned weights. These visualizations provide hints on which
gradients the learning found discriminative.  Notice the detailed structure
that emerges from our visualization that is not apparent in the HOG glyph. In
most cases, one can recognize the category of the detector by only looking at the
visualizations.

\section{Conclusion}


We believe visualizations can be a powerful tool for understanding object
detection systems and advancing research in computer vision. To this end, this
paper presented and evaluated four algorithms to visualize object detection
features. Since object detection researchers analyze HOG glyphs everyday and
nearly every recent object detection paper includes HOG visualizations, we hope
more intuitive visualizations will prove useful for the community.


{\footnotesize \paragraph{Acknowledgments:} We thank Hamed Pirsiavash, Joseph
Lim, the MIT CSAIL vision group, and anonymous reviewers. Funding for this
research was provided by a NSF GRFP to CV, a Facebook fellowship to AK, and a
Google research award, ONR MURI N000141010933 and NSF Career Award No. 0747120
to AT.

\bibliographystyle{ieee}
\bibliography{main}
}

\appendix
\section{Full Evaluation Tables}

\noindent We provide the full inversion benchmark with normalized cross correlation:
\begin{center}
\begin{tabular}{l | c c c c c}
Category & ELDA & Ridge & Direct & PairDict \\
\hline
aeroplane &      \textbf{0.634} &  \textbf{0.633} &  0.596 &  0.609 \\
bicycle &        0.452 &  \textbf{0.577} &  0.513 &  0.561 \\
bird &   \textbf{0.680} &  0.650 &  0.618 &  0.638 \\
boat &   \textbf{0.697} &  0.678 &  0.631 &  0.629 \\
bottle &         \textbf{0.697} &  0.683 &  0.660 &  0.671 \\
bus &    0.627 &  \textbf{0.632} &  0.587 &  0.585 \\
car &    0.668 &  \textbf{0.677} &  0.652 &  0.639 \\
cat &    \textbf{0.749} &  0.712 &  0.687 &  0.705 \\
chair &  \textbf{0.660} &  0.621 &  0.604 &  0.617 \\
cow &    \textbf{0.720} &  0.663 &  0.632 &  0.650 \\
table &    \textbf{0.656} &  0.617 &  0.582 &  0.614 \\
dog &    \textbf{0.717} &  0.676 &  0.638 &  0.667 \\
horse &  \textbf{0.686} &  0.633 &  0.586 &  0.635 \\
motorbike &      0.573 &  \textbf{0.617} &  0.549 &  0.592 \\
person &         \textbf{0.696} &  0.667 &  0.646 &  0.646 \\
pottedplant &    \textbf{0.674} &  \textbf{0.679} &  0.629 &  0.649 \\
sheep &  \textbf{0.743} &  0.731 &  0.692 &  0.695 \\
sofa &   \textbf{0.691} &  0.657 &  0.633 &  0.657 \\
train &  \textbf{0.697} &  0.684 &  0.634 &  0.645 \\
tvmonitor &      \textbf{0.711} &  0.640 &  0.638 &  0.629 \\
\hline
Mean & \textbf{0.671} & 0.656  &0.620 & 0.637\\
\end{tabular}
\end{center}

\noindent We provide the full visualization benchmark with human subjects:
\begin{center}
\begin{tabular}{l | c c c c c | c}
Category & ELDA & Ridge & Direct & PairDict & Glyph & Expert \\
\hline
aeroplane & 0.433& 0.391& 0.568& \textbf{0.645}& 0.297 & 0.333\\
bicycle & 0.327& 0.127& 0.362& 0.307& \textbf{0.405} & 0.438 \\
bird & 0.364& 0.263& \textbf{0.378}& 0.372& 0.193 & 0.059\\
boat & 0.292& 0.182& 0.255& \textbf{0.329}& 0.119 & 0.352\\
bottle & 0.269& 0.282& 0.283& \textbf{0.446}& 0.312 & 0.222\\
bus & 0.473& 0.395& \textbf{0.541}& 0\textbf{.549}& 0.122 & 0.118\\
car & 0.397& 0.457& \textbf{0.617}& 0.585& 0.359 & 0.389\\
cat & 0.219& 0.178& \textbf{0.381}& 0.199& 0.139 & 0.286 \\
chair & 0.099& 0.239& 0.223& \textbf{0.386}& 0.119 & 0.167\\
cow & 0.133& 0.103& \textbf{0.230}& 0.197& 0.072 & 0.214\\
table & 0.152& 0.064& 0.162& \textbf{0.237}& 0.071 & 0.125\\
dog & 0.222& 0.316& \textbf{0.351}& 0.343& 0.107 & 0.150\\
horse & 0.260& 0.290& 0.354& \textbf{0.446}& 0.144 & 0.150\\
motorbike & 0.221& 0.232& \textbf{0.396}& 0.224& 0.298 & 0.350\\
person & 0.458& 0.546& 0.502& \textbf{0.676}& 0.301 & 0.375\\
pottedplant & 0.112& 0.109& \textbf{0.203}& 0.091& 0.080 & 0.136\\
sheep & 0.227& 0.194& \textbf{0.368}& 0.253& 0.041 & 0.000\\
sofa & 0.138& 0.100& 0.162& \textbf{0.293}& 0.104 & 0.000\\
train & 0.311& 0.244& 0.316& \textbf{0.404}& 0.173 & 0.133\\
tvmonitor & 0.537& 0.439& 0.449& \textbf{0.682}& 0.354 & 0.666\\
\hline
Mean & 0.282& 0.258& 0.355& \textbf{0.383} & 0.191 & 0.233 
\end{tabular}
\end{center}

\vfill\break

\section{Visualization Class Confusions}

\noindent We show the confusion matrices for humans classifying our HOG inversions for various algorithms:
\begin{center}
\noindent\includegraphics[width=0.5\linewidth]{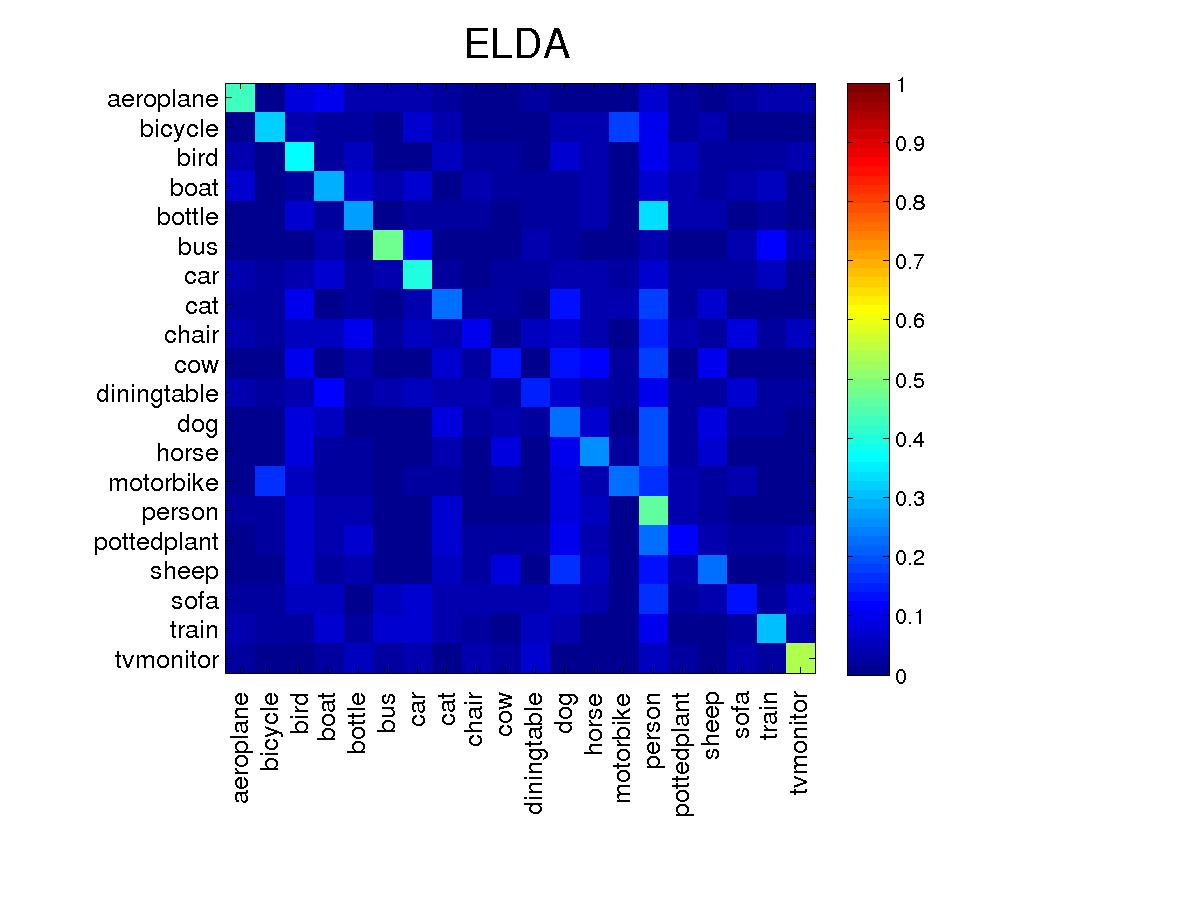}\includegraphics[width=0.5\linewidth]{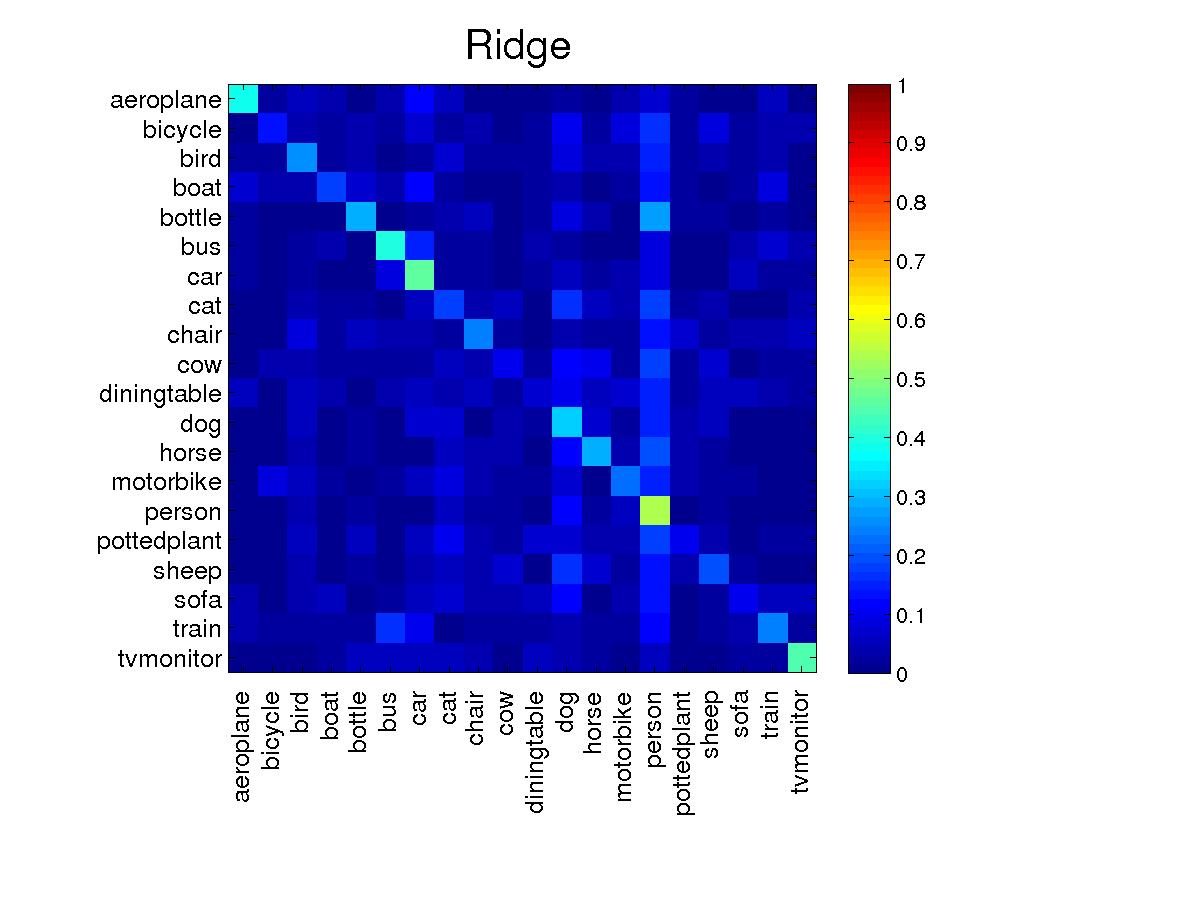}
\includegraphics[width=0.5\linewidth]{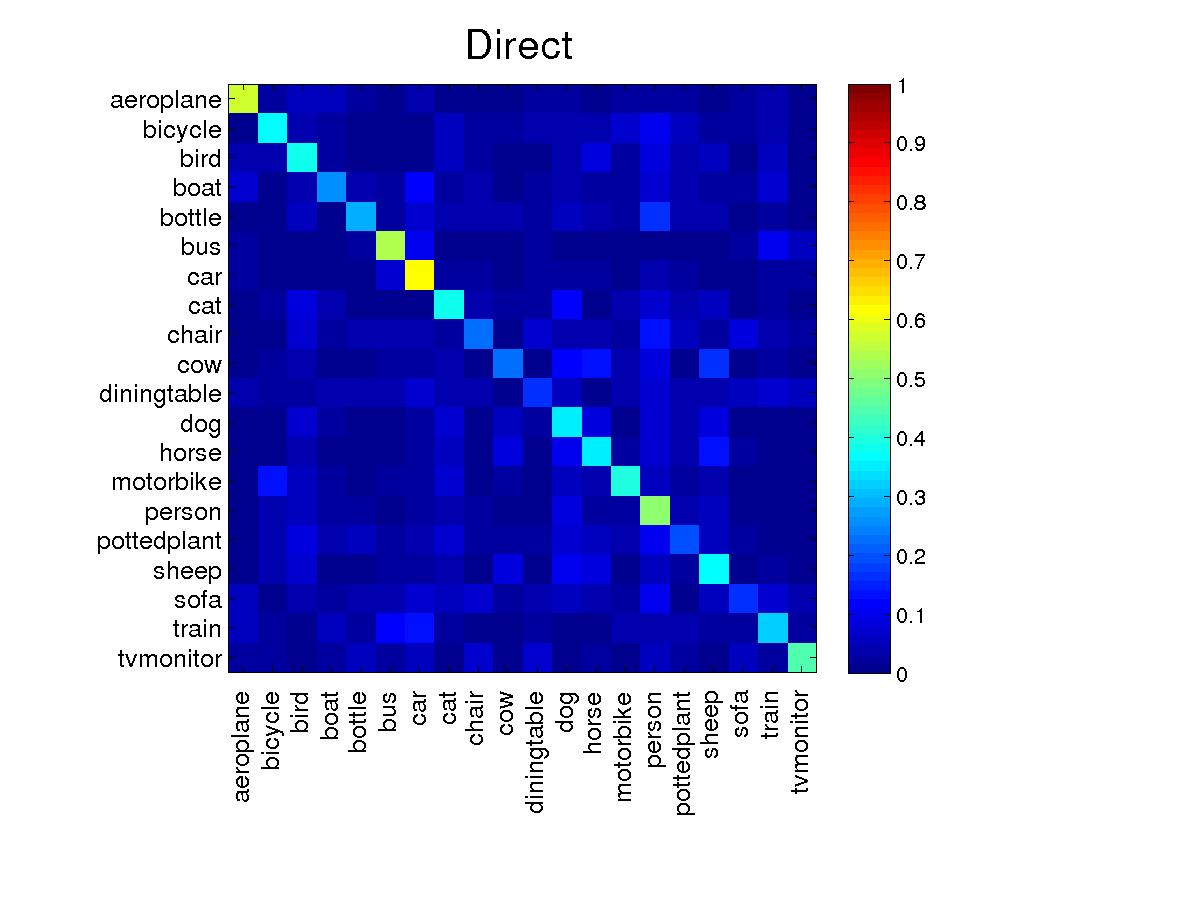}\includegraphics[width=0.5\linewidth]{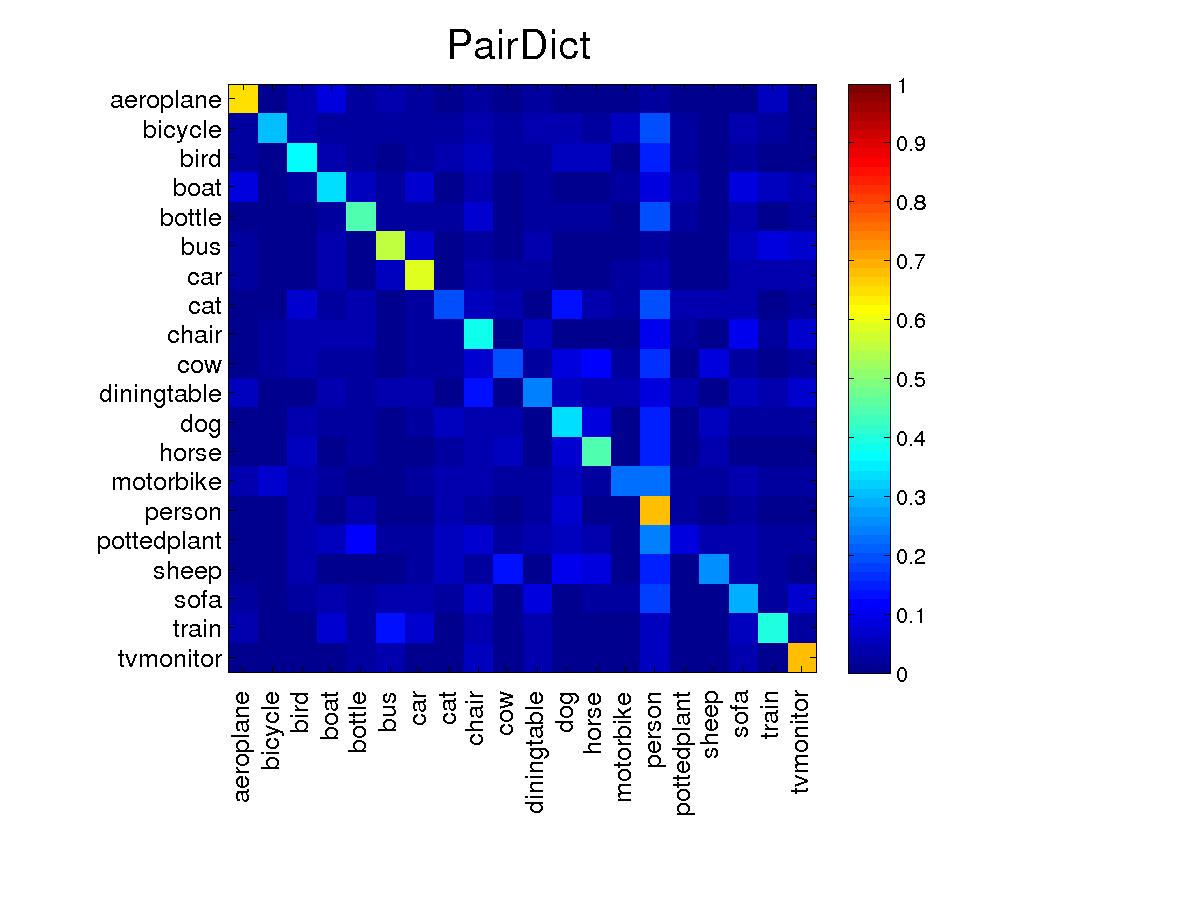}
\includegraphics[width=0.5\linewidth]{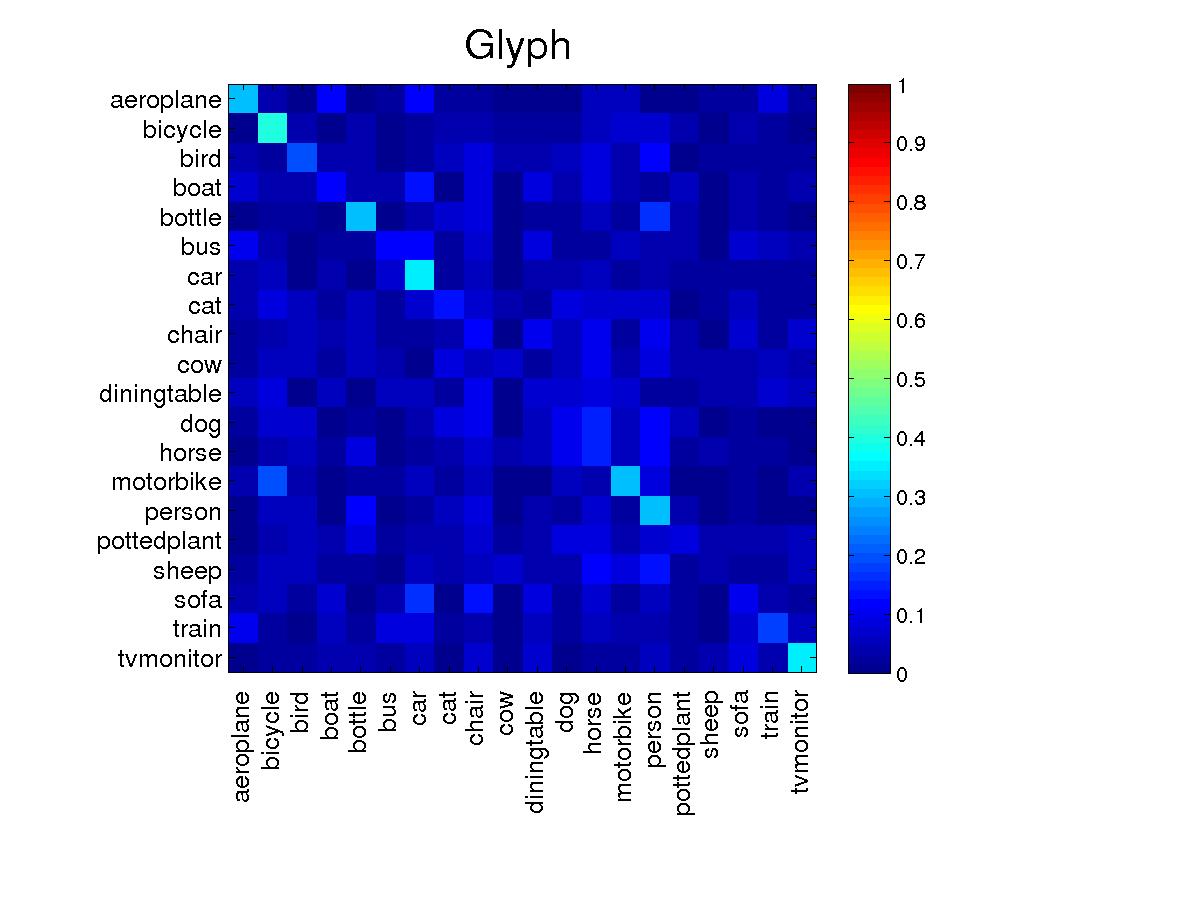}\includegraphics[width=0.5\linewidth]{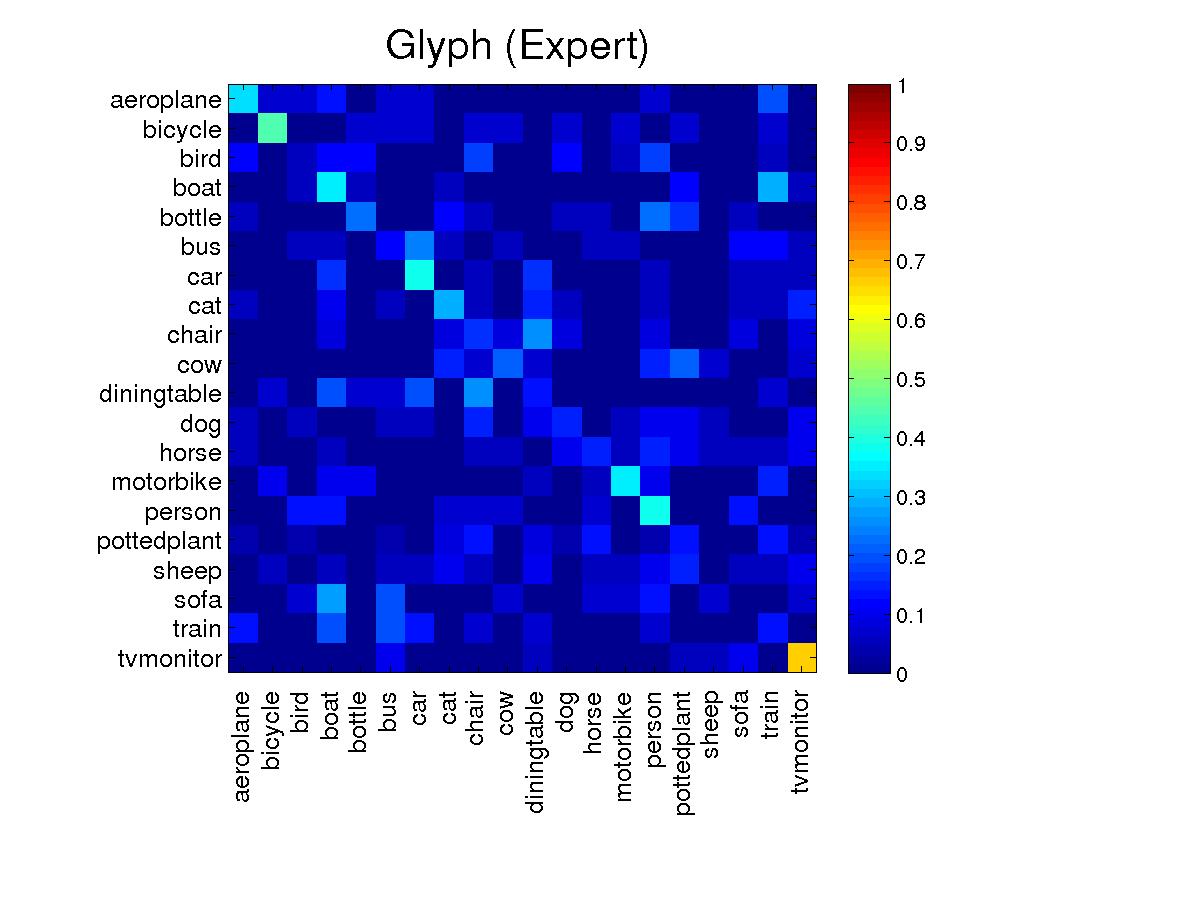}
\end{center}

\section{Additional Color Inversions}

\noindent On the left, we show the color inverse. On the right, we show the original 
image.

\begin{center}
\includegraphics[width=0.45\linewidth]{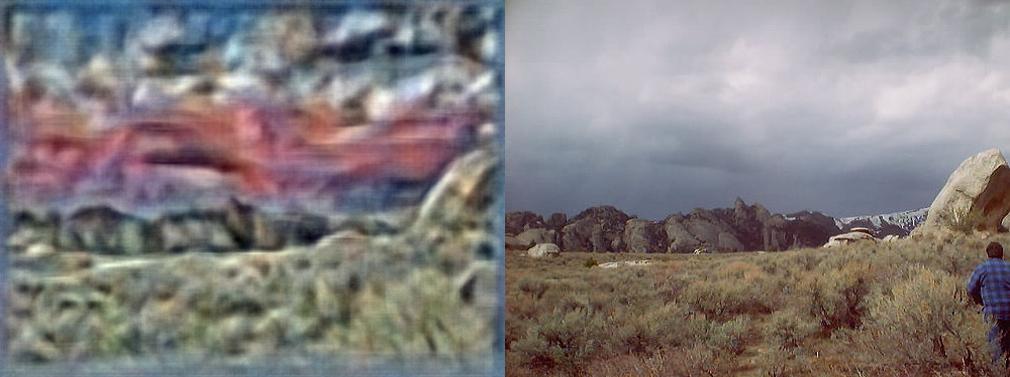}
\includegraphics[width=0.45\linewidth]{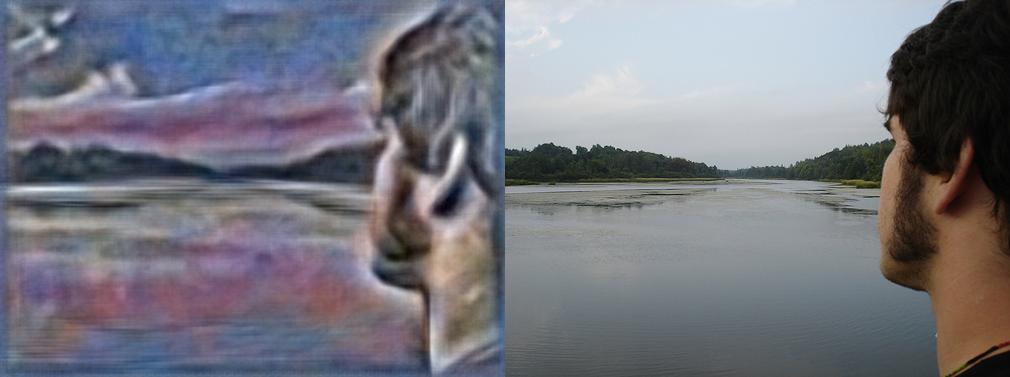}
\includegraphics[width=0.45\linewidth]{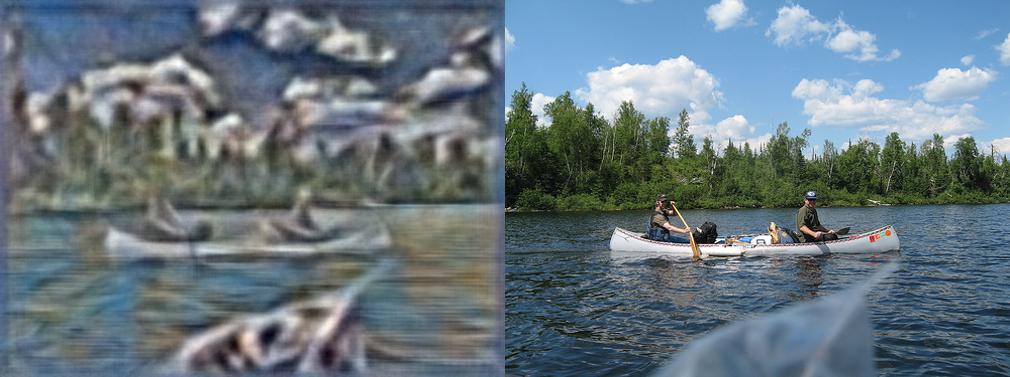}
\includegraphics[width=0.45\linewidth]{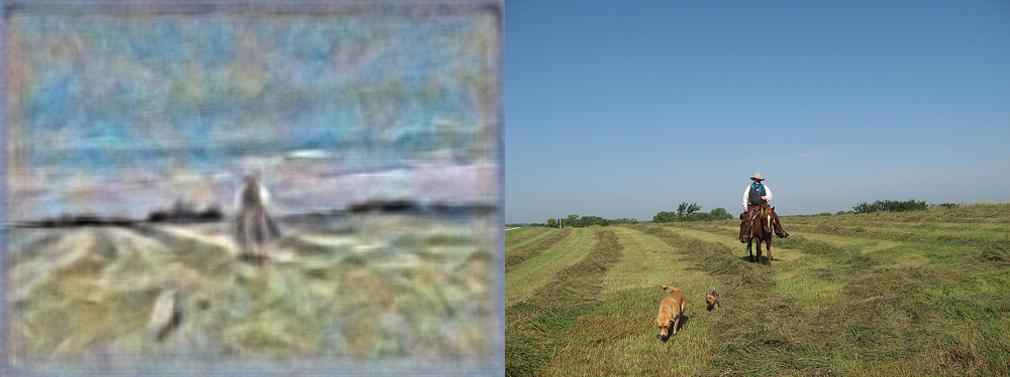}
\includegraphics[width=0.45\linewidth]{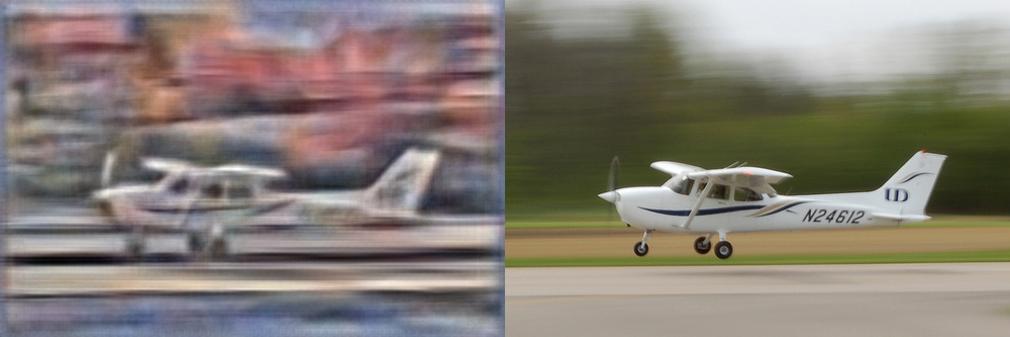}
\includegraphics[width=0.45\linewidth]{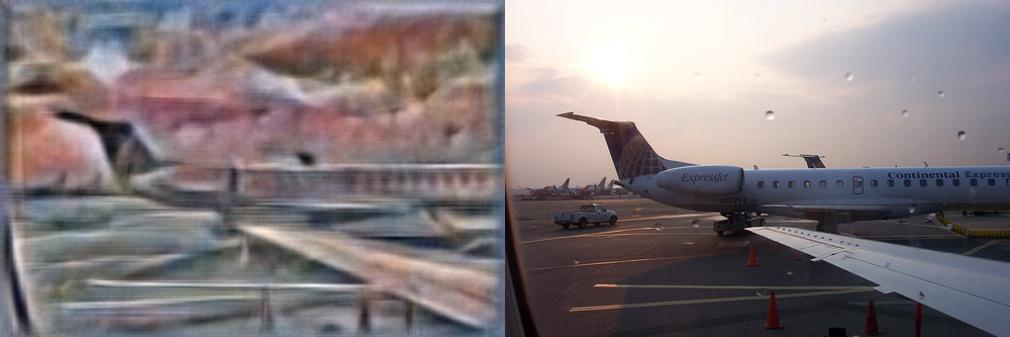}
\includegraphics[width=0.45\linewidth]{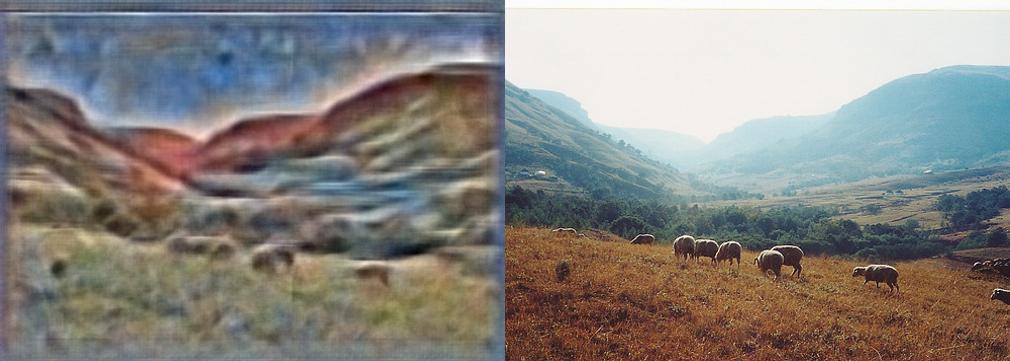}
\includegraphics[width=0.45\linewidth]{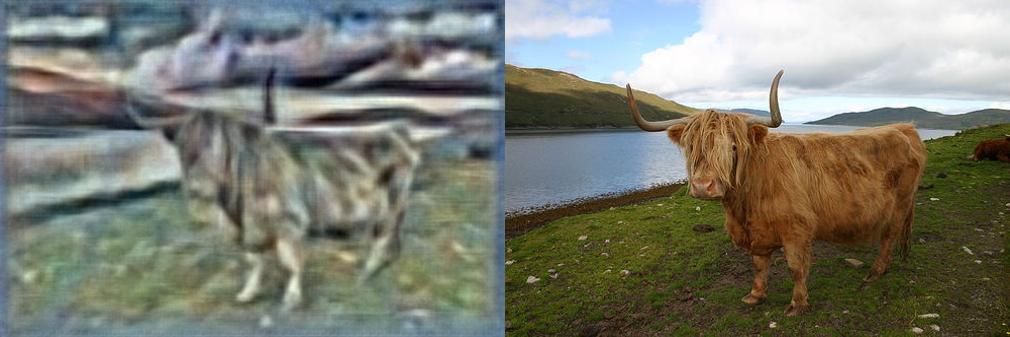}
\includegraphics[width=0.45\linewidth]{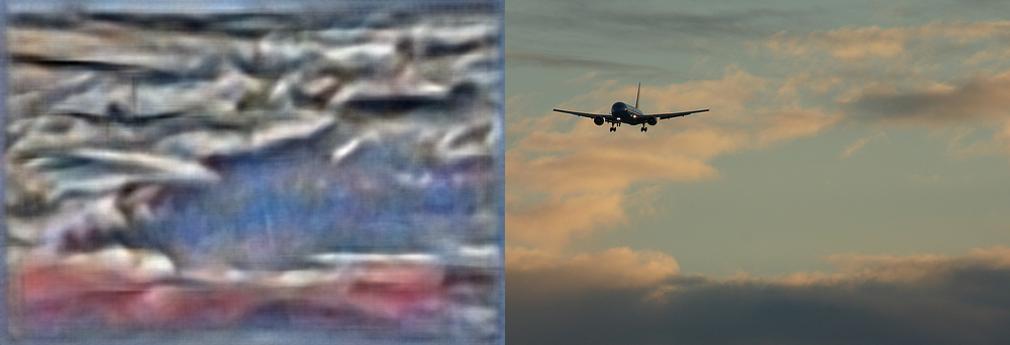}
\includegraphics[width=0.45\linewidth]{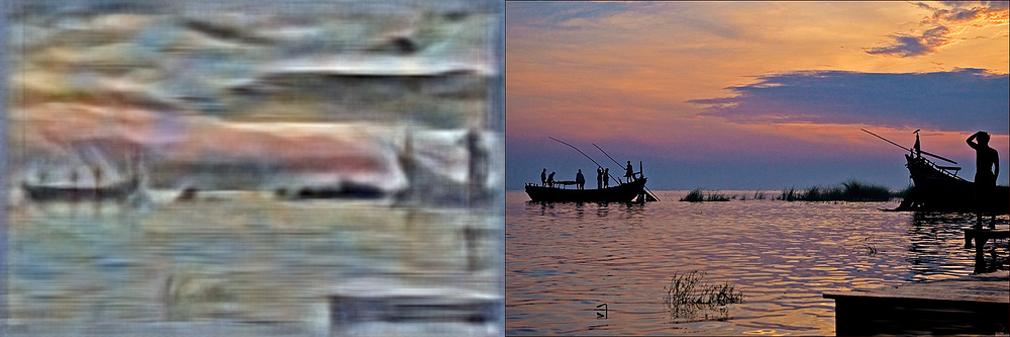}
\includegraphics[width=0.45\linewidth]{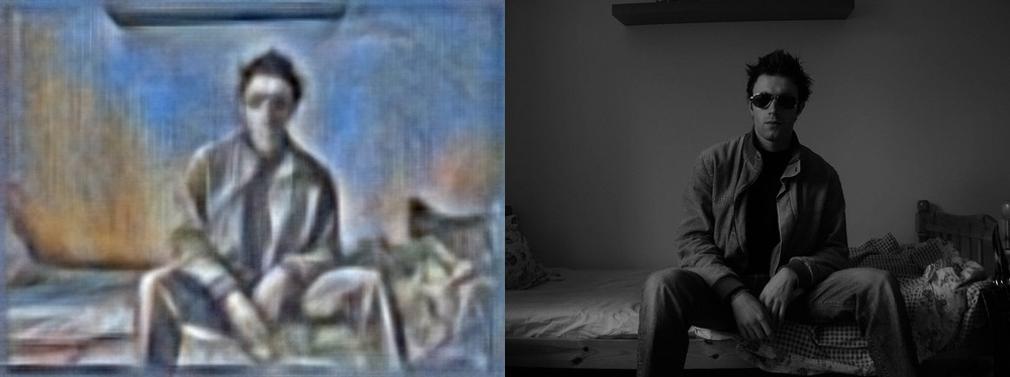}
\includegraphics[width=0.45\linewidth]{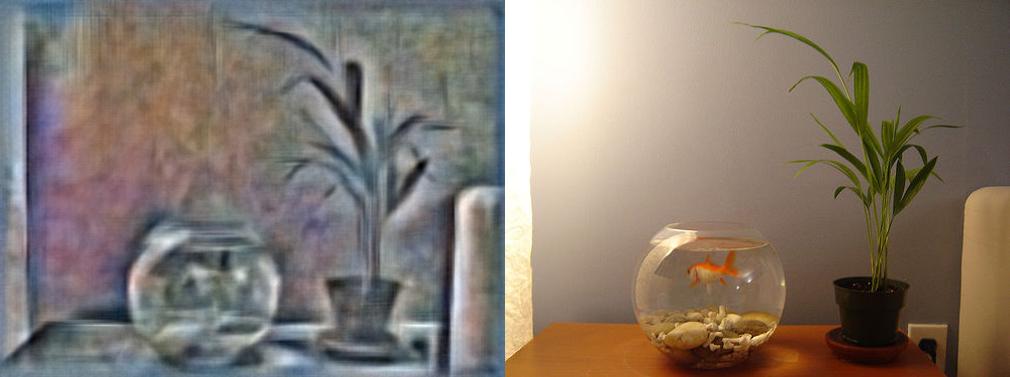}
\end{center}

\vfill

\end{document}